\documentclass{article}
\usepackage{preprint,times}

\usepackage{amsmath,amsfonts,bm}

\def\eqref#1{equation~\ref{#1}}

\def\1{\bm{1}}

\DeclareMathAlphabet{\mathsfit}{\encodingdefault}{\sfdefault}{m}{sl}
\SetMathAlphabet{\mathsfit}{bold}{\encodingdefault}{\sfdefault}{bx}{n}

\usepackage{float}
\usepackage{hyperref}
\usepackage{url}
\usepackage{amsmath,amssymb,booktabs,graphicx,hyperref,xcolor}
\usepackage{colortbl,tabularx}
\definecolor{policyikrow}{RGB}{228,237,244}
\newcommand{\sd}[1]{{\scriptsize\color{gray}$\pm$#1}}

\usepackage[normalem]{ulem}
\usepackage{enumitem}
\usepackage{placeins}
\usepackage{wrapfig}
\usepackage{dsfont}
\usepackage{etoc}
\makeatletter\providecommand\captionof[1]{\def\@captype{#1}\caption}\makeatother

\title{PolicyIK: High-Precision Inverse Kinematics for Robot Arms with Reinforcement Learning}

\author{Yuhan Liu$^{1*}$, Xinqi Liu$^{1*}$, Zixuan Dong$^{1*}$, Keith W. Ross$^{2}$, Shenji Wan$^{1}$ \\
$^{1}$New York University \quad $^{2}$New York University Abu Dhabi \\
\texttt{\{yl10379, xl4600, zixuandong, keithwross, sw5973\}@nyu.edu} \\
$^{*}$Equal contribution
}

\begin{document}

\maketitle
\lhead{Preprint}

\begin{abstract}
Inverse kinematics (IK) converts end-effector pose commands into joint motion, and many vision-language-action (VLA) policies rely on it to act on a given arm and to transfer across embodiments. Precise IK remains laborious: redundancy and singularities complicate the inverse mapping, execution depends on hand-tuned controllers, and training data on physical robots are costly to collect. We present PolicyIK, a goal-conditioned reinforcement learning policy for high-precision IK. It updates joint-position references from pose feedback and observes its accumulated reference relative to the measured joints, so it can retain the offset that the position controller needs at rest. Two-scale pose errors with an error-dependent action gain resolve millimeter residuals while preserving large motions. Progress and dwell rewards teach it to both reach the tolerance and stay within it. In simulation, a single attempt by the trained PPO policy keeps the mean errors over the final second within 1\,mm and $0.2^\circ$ on 93.5\% of held-out tasks, and a deployment protocol raises success to 98.9\%. When trained directly on a physical arm with an off-policy RL learner, PolicyIK achieves higher reaching and holding success than all four classical IK controllers under the same time budgets. In simulated pick-and-place, PolicyIK executes the end-effector commands of the $\pi_{0.5}$ VLA policy with 95.7\% success. These results show that reinforcement learning can provide high-precision inverse kinematics for robot arms. Code is available at \url{https://github.com/Tinaliu0123/PolicyIK}.

\end{abstract}

\etocdepthtag.toc{mainbody}
\section{Introduction}
\label{sec:intro}
Many learned manipulation policies act on a robot arm through inverse
kinematics (IK), which turns end-effector pose commands into joint motion.
Vision-language-action (VLA) and generalist policies often predict
end-effector poses or pose increments and leave the conversion to joint
commands to an IK or operational-space controller
\citep{oxe2024,octo2024,openvla2024}. Because this interface is shared across
arms, the same policy can run on different robots
\citep{chi2024umi,mirage2024,shadow2024,umi2024}. Such transfer delegates the
feasibility and accuracy of every command to the target robot's IK. Transfer
methods assume controllers that reach commanded poses within a few millimeters
\citep{mirage2024}, yet transfer fails at joint limits and on motions the arm cannot follow \citep{chi2024umi,polybot2023}. IK accuracy thus
carries over to every policy that acts through it.

Developing and deploying a precise pose controller raises three challenges. First, the inverse
mapping itself is difficult. A redundant arm can reach a pose with many joint configurations \citep{burdick1989}, and near singularities small pose corrections demand large joint motions, so numerical solvers trade accuracy for robustness and can fail even on reachable poses \citep{buss2004,deo1995,beeson2015}.
Second, the realized pose depends on robot-specific execution. Finite gains,
gravity, and joint limits shift where a position-controlled arm comes to rest. Controller dynamics also affect
how action errors propagate during policy execution~\citep{bronars2026tune}.
The controllers that compensate for these effects rely on damping factors, task
weights, and gains that experts tune for each robot
\citep{lynch2017,charbonneau2018,buss2005,lee2026space}. Third, collecting
training data on physical robots is costly, as each trial requires robot
execution and a reset.

We investigate reinforcement learning (RL) for this setting. An RL policy learns the
mapping from pose error to joint commands through interaction with the robot's
dynamics. Sampling goals from reachable configurations yields feasible goals
across the workspace, and the designed reward targets the required tolerance. Prior RL controllers have reached position goals with
fixed-base arms and tracked full poses with legged manipulators, with errors
from several millimeters to a few centimeters
\citep{jailer2021,aljalbout2024,portela2025,pake2026}, and a hybrid reaches
sub-millimeter accuracy in kinematic simulation by pairing the policy with
damped least squares, which supplies the final correction \citep{mpdik2025}.
Reaching millimeter-level full-pose accuracy with the policy alone calls for a
formulation designed around terminal precision.

\begin{figure}[t]
\centering
\includegraphics[width=\linewidth]{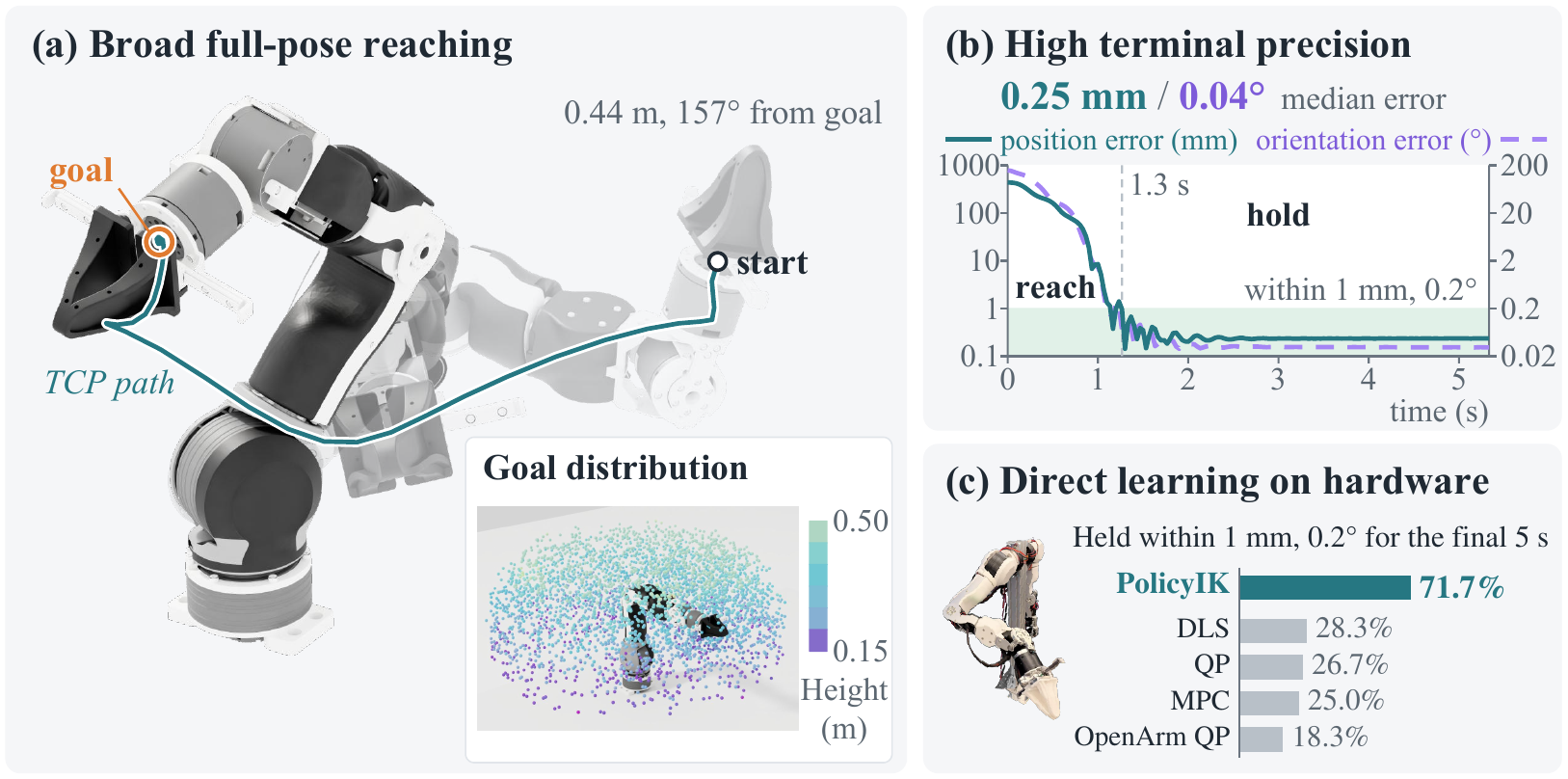}
\caption{\textbf{PolicyIK: from broad reaching to precise regulation.} (a) Full-pose reaching across the simulated workspace. (b) Reaching and holding within 1\,mm and $0.2^\circ$. (c) On hardware, PolicyIK holds the pose within 1\,mm and $0.2^\circ$ throughout the final 5\,s on more tasks than classical controllers.} 
\label{fig:task-overview}
\vspace{-0.5cm}
\end{figure}

We present PolicyIK, a goal-conditioned RL formulation that learns
high-precision IK for robot arms through a joint-position interface
(Figure~\ref{fig:control-loop}). Like closed-loop IK \citep{siciliano1990}, the
policy updates a joint reference from pose feedback, but learns this update
from interaction. The formulation addresses three properties of precise
regulation. First, the policy accumulates its actions in the reference and
observes the reference relative to the measured joints, so it can retain the
offset that the position controller needs at rest. Second, pose errors enter
the observation at coarse and fine scales, and an error-dependent gain shrinks
actions near the goal, which resolves small residuals while preserving large
motions. Third, the reward credits progress beyond the best error so far and time within tolerance, so it trains both reaching and holding.
We train with PPO \citep{schulman2017ppo} in parallel simulation
\citep{orbit2023} and with FlashSAC \citep{kim2026flashsac} on physical arms.
Since every hardware transition costs robot time, a reverse curriculum and
near-goal sampling focus training on the terminal phase, while a
latent-alignment critic handles large-rotation detours.

We evaluate PolicyIK in simulation and on a physical arm (Figure~\ref{fig:task-overview}). On a
simulated 7-DoF arm, PolicyIK reaches median errors of 0.25\,mm and
$0.04^\circ$ and meets the 1\,mm and $0.2^\circ$ tolerance on 93.5\% of held-out
tasks in one attempt and on 98.9\% with a deployment protocol. Removing
any design component lowers single-attempt success by 1.8--9.2 points.
Direct hardware training demonstrates the value of learning to maintain a pose
after reaching it. PolicyIK achieves higher reaching and holding success than all four
classical IK controllers under the same time budgets. In simulated pick-and-place, it executes the
end-effector commands of $\pi_{0.5}$~\citep{pi05_2025} with 95.7\% success
across ten adapted LIBERO-Object tasks~\citep{libero2023}, without pick-and-place data in its training.
These capabilities come with a low inference cost: an actor forward pass takes
0.035\,ms on CPU, well below the 33\,ms control period.
Our contributions are as follows:
\begin{itemize}[leftmargin=*,itemsep=2pt,parsep=0pt,topsep=3pt]
\item We show that goal-conditioned RL formulates inverse kinematics for robot arms and can train a policy to reach and hold millimeter-level full-pose
precision through a joint-position interface.
\item We identify three design choices behind this precision: observing the
joint reference relative to the measured joints, sensing pose errors at
centimeter and sub-millimeter scales with smaller steps near the goal, and
rewarding each new best error and every step within tolerance. Ablations show
they affect reaching and holding separately.
\item We train PolicyIK directly on two physical arms (OpenArm and xArm7) with
FlashSAC \citep{kim2026flashsac}, hindsight goal relabeling
\citep{andrychowicz2017hindsight}, and a curriculum from near-goal starts
\citep{pmlr-v78-florensa17reverse} to large rotations. It holds reached poses
within 1 mm and $0.2^\circ$ more reliably than classical IK controllers, and executes a
VLA policy's commands in simulated pick-and-place with 95.7\% success.
\end{itemize}

\section{Related Work}
\vspace{-0.05cm}
\label{sec:related}
\paragraph{Inverse kinematics and task-space control.}
Classical approaches map pose commands to joint motion analytically,
numerically, or through feedback. Closed-form solutions exist mainly for arms
with special geometry \citep{diankov2010}, so general and redundant
arms rely on damped least squares or constrained optimization
\citep{wampler1986,deo1995,beeson2015}, and closed-loop controllers convert
task-space error into joint references under joint limits and secondary
objectives \citep{siciliano1990,haviland2023,rmpflow2021,fabrics2021}. These
methods require tuning for each robot \citep{charbonneau2018,buss2005} and
leave execution of joint motion to the robot's controller. Learned IK models fit the inverse map
from sampled configurations or demonstrations
\citep{ikflow2022,habekost2024,mimicik2026} and rely on numerical refinement for
high accuracy \citep{cppflow2024}.
In contrast, PolicyIK learns closed-loop IK from interaction with the robot's
dynamics and regulates the realized pose.

\vspace{-0.15cm}
\paragraph{Reinforcement learning for reaching and pose tracking.}
Reinforcement learning controllers for end-effector reaching range from
benchmark tasks to fixed-base arms and legged manipulators. Common multi-goal
benchmarks count a reach as successful within 5\,cm of a position goal
\citep{plappert2018,metaworld2019}. Policies for fixed-base arms map
task-space errors to joint commands and reach position goals with errors of
millimeters to centimeters \citep{jailer2021,aljalbout2024}, while whole-body
controllers track full end-effector poses on legged manipulators to within a
few centimeters \citep{umi2024,portela2025,pake2026}. When tighter tolerances
are needed, learning is paired with a classical solver, such as damped least
squares, that performs the final correction \citep{mpdik2025}. PolicyIK learns joint-reference updates for millimeter-level full-pose reaching and sustained holding, without a numerical IK solver.

\vspace{-0.15cm}
\paragraph{Learning for execution-error compensation.}
Residual methods add learned corrections to a conventional controller or its
reference trajectory \citep{residual2019,pavlichenko2022} and build on its
tuning. Integrating methods accumulate policy outputs in the commanded state
to reduce steady-state error \citep{industreal2023,aljalbout2024}. Integral
state augmentation exposes accumulated information for reference tracking
\citep{iasa2022}. Once commands are integrated, the accumulated
reference becomes part of the system state. PolicyIK observes this reference
relative to the measured joints and pairs it with two-scale pose errors and
error-dependent action scaling, so one policy handles both large motions and
millimeter corrections.

\section{Problem Formulation}
\vspace{-0.15cm}
\label{sec:problem}
Classical IK computes a joint configuration from a goal end-effector pose, but commanding that joint target on a position-controlled arm
does not guarantee that the end effector reaches the target pose: gravity, finite gains,
and joint limits shift where the arm comes to rest. The controller must therefore close the loop on
the pose error, drive it below the tolerance, and hold it there. We formulate this closed-loop regulation as a goal-conditioned RL problem. 

A goal-conditioned Markov
Decision Process (MDP) is a tuple
$(\mathcal{S},\mathcal{A},\mathcal{G},P,r,\nu_0,\gamma)$, in which a policy
$\pi(a\mid s,g)$ chooses actions $a\in\mathcal{A}$ in states $s\in\mathcal{S}$
to reach a goal $g\in\mathcal{G}$. The transition $P(s'\mid s,a)$ is shared
across goals, and the policy maximizes the expected sum of rewards $r(s,a,g)$
discounted by $\gamma$, starting from $(s_0,g)\sim\nu_0$.
For a 7-DoF arm in our task:

\begin{itemize}[leftmargin=*,itemsep=2pt,parsep=0pt,topsep=3pt]
\item The \emph{goal} is a pose $g=T_g=(R_g,p_g)\in SE(3)$, the group of rigid-body
poses with rotation $R_g\in SO(3)$ and position $p_g\in\mathbb{R}^3$.
\item The \emph{state} $s_t=(q_t,\dot q_t,z_{t-1},a_{t-1})$ contains the measured
joint positions $q_t\in\mathbb{R}^7$ and velocities $\dot q_t$, the previous
joint reference $z_{t-1}$, and the previous action. The end-effector pose
follows from forward kinematics, $T_t=(R_t,p_t)=\operatorname{FK}(q_t)$.
\item The \emph{action} $a_t\in\mathbb{R}^7$ is an increment to the joint reference
$z_t$, which will be formally introduced later in Equation~\ref{eq:reference-update}.
\item The \emph{transition} $P$ tracks $z_t$ with a fixed-gain position controller for
one control period ($\Delta t=1/30$\,s in simulation) under the rigid-body dynamics and the
joint, command, and torque limits. Because the same action moves the arm
differently under different references, $z_{t-1}$ is part of the state.
\item The \emph{initial distribution} $\nu_0$ samples a collision-free start
configuration $q_0$ at rest with $z_{-1}=q_0$, and a goal
$T_g=\operatorname{FK}(q^*)$ at a reachable configuration $q^*$ inside the defined workspace (Appendix~\ref{app:mdp}).
\end{itemize}

\paragraph{Success criterion.}
Let $e_p(t)=\lVert p_t-p_g\rVert_2$ be the position error and
$e_R(t)=\lVert\operatorname{Log}(R_g^\top R_t)^\vee\rVert_2$ the rotation angle
between $R_t$ and $R_g$. The base reward is the sparse success indicator
\begin{equation}
r(s_t,a_t,g)=\mathds{1}\big[\,e_p(t)\le 1\,\mathrm{mm}\ \wedge\ e_R(t)\le 0.2^\circ\,\big],
\label{eq:success}
\end{equation}
which we maximize over an episode, ending only at the time limit.
Maximizing it therefore rewards both reaching the tolerance early and staying
within it. We evaluate with the same tolerances: an episode \emph{strictly succeeds}
when the mean errors over its final $H=30$ steps (1\,s) satisfy
Equation~\ref{eq:success}. Because the sparse reward is rarely observed from
distant starts, Section~\ref{sec:reward} will design dense reward terms. As diagnostics, the
\emph{coarse} criterion relaxes the tolerances to 1\,cm and $2^\circ$, about
the accuracy of prior learned reaching policies in the literature,
and \emph{containment} requires Equation~\ref{eq:success} at every step of
the final window rather than on average.

\section{Method}
\label{sec:method}
\begin{figure}[t]
\centering
\includegraphics[width=\linewidth]{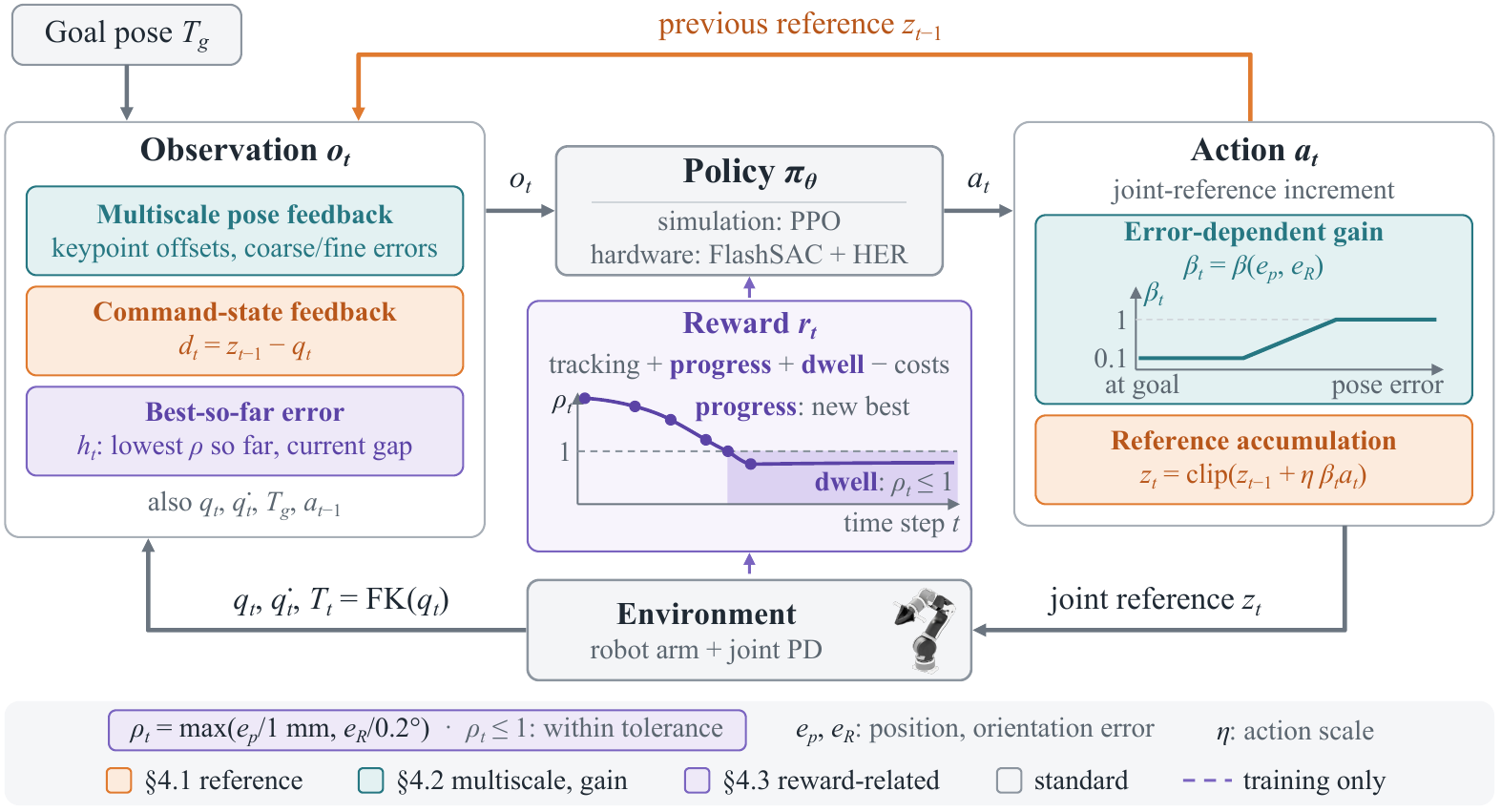}
\caption{\textbf{PolicyIK control formulation.} At each step, the policy maps the
observation to a joint-reference increment, which is scaled by the
error-dependent gain, accumulated in the joint reference, and tracked by a joint
PD controller. Colored blocks are the designs of
Sections~\ref{sec:observations} to~\ref{sec:reward}.}
\vspace{-1.3em}
\label{fig:control-loop}
\end{figure}

\vspace{-0.1cm}
At each step, PolicyIK updates a joint reference that the position
controller tracks (Figure~\ref{fig:control-loop}).
Holding a pose to a millimeter through this interface raises three problems
that coarse reaching can tolerate: the command that holds a pose is state that
the joint measurements do not reveal (Section~\ref{sec:observations}), the
residuals that decide success lie orders of magnitude below the initial errors
(Section~\ref{sec:precision-feedback}), and dense tracking rewards barely
distinguish meeting the tolerance from missing it (Section~\ref{sec:reward}). Training directly on a physical arm further limits the interaction budget, which Section~\ref{sec:real-implementation} addresses with adaptations to the real-world deployment.

\subsection{Reference Accumulation and Command-State Feedback}
\label{sec:observations}
\label{sec:actions}
Under gravity and finite gains, an arm settles away from its
reference, so holding a goal requires a reference offset from the goal
configuration (Appendix~\ref{app:mdp}). If actions are applied relative to the measured joints, the policy must regenerate this offset with nonzero actions at every step, and the arm sags whenever actions relax. PolicyIK
instead accumulates its actions in the reference,
\begin{equation}
z_t=\operatorname{clip}\!\left[z_{t-1}+\eta\,\beta(e_p,e_R)\,a_t,\;
q^{\mathrm{lo}},\;q^{\mathrm{hi}}\right],
\label{eq:reference-update}
\end{equation}
so that a zero action retains the offset, a multi-step integration in the
terminology of \citet{aljalbout2024}. Here $\eta$ is the action scale, $\beta$ is the
error-dependent gain of Section~\ref{sec:precision-feedback}, and
$q^{\mathrm{lo}},q^{\mathrm{hi}}$ are the joint limits.

Accumulation makes the reference part of the state: the same joint state, goal,
and action lead to different next states under different references. Without
observing the reference, the policy would have to infer it from its action
history. PolicyIK therefore feeds back the command state $d_t=z_{t-1}-q_t$, the
offset between the reference and the measured joints, within the augmented policy input $o_t$, built from $s_t$ and $g$
\begin{equation}
o_t=\operatorname{concat}\bigl(q_t-\bar q,\,\dot q_t,\,p_g,\,
\Delta k_t,\,v_t^{c},\,v_t^{f},\,h_t,\,d_t,\,a_{t-1}\bigr),
\label{eq:observation}
\end{equation}
where $\bar q$ is a nominal configuration, poses are in base frame,
$\Delta k_t$, $v_t^{c}$,
and $v_t^{f}$ are pose features (Section~\ref{sec:precision-feedback}),
and $h_t$ exposes running-best error (Section~\ref{sec:reward}). The actor
and critic share this observation.

\subsection{Multiscale Pose Feedback and Action Scaling}
\label{sec:precision-feedback}
Correcting millimeter residuals requires the policy both to perceive them and to
resolve them with actions. Without error channels scaled to the tolerance, we
observed the arm settling at a biased equilibrium short of the goal. Input normalization scales each channel by its spread over the workspace,
so millimeter differences occupy a negligible fraction of the joint,
goal, and keypoint inputs. PolicyIK therefore represents the pose error at two
scales and adapts its step size to the error.

\paragraph{Multiscale pose feedback.}
Following keypoint-based pose representations \citep{portela2025}, three
noncollinear local offsets $c_i$ define
\begin{equation}
\Delta k_{i,t}=(p_g+R_gc_i)-(p_t+R_tc_i),
\label{eq:keypoints}
\end{equation}
which express translation and orientation errors as position differences.
To let the policy perceive millimeter errors, we also express the error in the
current end-effector frame, clipped at a coarse and a fine scale,
\begin{equation}
v_t^{\ell}=\begin{bmatrix}
\operatorname{clip}\bigl(R_t^\top(p_g-p_t)/\delta_\ell,-1,1\bigr)\\
\operatorname{clip}\bigl(\operatorname{Log}(R_t^\top R_g)^\vee/\alpha_\ell,-1,1\bigr)
\end{bmatrix},\qquad \ell\in\{c,f\},
\label{eq:pose-channels}
\end{equation}
with $(\delta_c,\alpha_c)=(1\,\mathrm{cm},20^\circ)$ and
$(\delta_f,\alpha_f)=(0.5\,\mathrm{mm},0.25^\circ)$. The coarse channels resolve errors
below 1\,cm and $20^\circ$, and the fine channels resolve errors at the scale of the tolerance: an
orientation error of $0.2^\circ$ fills 1\% of the coarse range but 80\% of the
fine range.

\paragraph{Error-dependent action scaling.}
A single action scale sets both the speed of transport and the resolution of
the final correction, because the reference moves by $\eta$ times any error in the
action. With a fixed scale, small values slowed transport and large values lowered terminal precision. PolicyIK therefore scales each update
by the gain
\begin{equation}
\begin{split}
\beta(e_p,e_R)&=\max\{G(e_p;l_p,u_p),G(e_R;l_R,u_R)\},\\
G(e;l,u)&=\beta_{\min}+(1-\beta_{\min})
\operatorname{clip}\!\left(\frac{e-l}{u-l},0,1\right),
\end{split}
\label{eq:action-gain}
\end{equation}
which keeps the full step while either error exceeds its upper threshold $u$ and
reduces it to $\beta_{\min}\eta$ once both errors fall below their lower thresholds $l$
(Appendix~\ref{app:mdp}). Because the gain follows the
error rather than time, fine resolution engages whenever the arm nears
the goal, regardless of duration.

\subsection{Reward Design}
\label{sec:reward}
The reward combines terms for the approach and for the terminal phase:
\begin{equation}
r_t=r_t^{\mathrm{track}}+r_t^{\mathrm{progress}}+r_t^{\mathrm{dwell}}
-c_t^{\mathrm{motion}}-c_t^{\mathrm{limit}}.
\label{eq:total-reward}
\end{equation}
Tracking guides the approach, progress and dwell reward reaching and holding the
tolerance, and the costs regularize motion. Appendix~\ref{app:training} lists
all terms and weights.

\paragraph{Tracking.}
The tracking reward
\begin{equation}
r_t^{\mathrm{track}}=-0.2\,e_p(t)+0.1\,K(e_p(t);0.1)+K(\varepsilon_t;0.2)+K(\varepsilon_t;0.05),
\label{eq:tracking-reward}
\end{equation}
where $e_p(t)$ is in meters,
$\varepsilon_t$ is the mean of
\mbox{$\lVert(p_t+R_t\tilde c_i)-(p_g+R_g\tilde c_i)\rVert_2$} over three keypoint
offsets $\tilde c_i$ (Appendix~\ref{app:training}), and
\mbox{$K(x;\sigma)=1-\tanh(x/\sigma)$}. Each kernel
is bounded and steepest within its bandwidth $\sigma$, which strengthens the pull
to the goal without growing at large errors. Within 1\,mm of the
goal, each kernel changes by at most 2\% of its range, a weak signal at
the tolerance scale.

\paragraph{Tolerance-normalized error.}
Success requires the position and orientation errors to satisfy their
tolerances at once. A
reward on summed errors lets progress on one error compensate for the other,
because the sum rewards reducing whichever error is easier while the other
remains outside its tolerance. The precision
terms therefore use the tolerance-normalized error and its running minimum,
\begin{equation}
\rho_t=\max\Bigl\{\frac{e_p(t)}{1\,\mathrm{mm}},\frac{e_R(t)}{0.2^\circ}\Bigr\},
\qquad b_t=\min\{b_{t-1},\rho_t\}.
\label{eq:rho}
\end{equation}
Thus $\rho_t\le1$ exactly when both current errors are at most 1\,mm and $0.2^\circ$, and $\rho_t$
decreases only when the binding error decreases. The running minimum $b_t$,
reset for each new goal, records the best accuracy reached so far. The
observation includes $h_t$, which contains $b_{t-1}$ and the gap
$\rho_t-b_{t-1}$ after clipping and scaling (Appendix~\ref{app:training}).

\paragraph{Progress.}
The progress reward drives $\rho_t$ below one with three terms of distinct
roles:
\begin{equation}
\begin{split}
r_t^{\mathrm{progress}}&=w_n r_t^{\mathrm{imp}}+r_t^{\mathrm{shape}}-w_b c_t^{\mathrm{back}},\qquad
r_t^{\mathrm{imp}}=\bigl[U(\rho_t)-U(b_{t-1})\bigr]_+,\\
r_t^{\mathrm{shape}}&=\gamma\Phi_t-\Phi_{t-1},\qquad
c_t^{\mathrm{back}}=\min\{3,[\rho_t-b_{t-1}]_+\},
\end{split}
\label{eq:progress-reward}
\end{equation}
where $[x]_+=\max(x,0)$, $U(\rho)=\tfrac12[1+\tanh(2(1-\rho))]$ rises from
0 to 1 as $\rho$ crosses one, $\Phi_t=-\min(50,\rho_t)$, and $w_n$ and $w_b$ are
weights. The improvement term
$r_t^{\mathrm{imp}}$ pays only when the running best improves. Because $b_t$
never increases, these payments telescope, so each accuracy level is paid once
and staying earns nothing. The term can therefore carry a large weight without
rewarding oscillation or stopping short of the tolerance, and $U$ concentrates
its payment on crossing into the tolerance. The potential-based shaping term $r_t^{\mathrm{shape}}$ \citep{ng1999policy} adds a dense
signal within 50 tolerance units, which also rewards returning toward a best
that the improvement term no longer pays for. The backslide cost $c_t^{\mathrm{back}}$ charges losses
relative to the best and discourages drift away from the accuracy already
reached. Its cap of three tolerance units lets it grow with small drifts while
staying constant for the larger excursions that reconfiguration requires.

\paragraph{Dwell.}
The dwell reward $r_t^{\mathrm{dwell}}=\mathds{1}[\rho_t\le1]$, the success indicator of
Equation~\ref{eq:success}, pays for every step inside the tolerance. The improvement term pays nothing for staying, and
without the dwell reward, policies more often touched the tolerance and then left it. Dwell rewards retaining accuracy, while the
improvement term rewards acquiring it. For the same reason, episodes end only at
time-out, since ending a goal at success would forfeit the dwell rewards that
follow.

\paragraph{Costs.}
The motion cost $c_t^{\mathrm{motion}}$ penalizes squared action rates and joint
velocities. The limit cost $c_t^{\mathrm{limit}}$ keeps the worst joint away from its limits and applies only while
$\rho_t>10$, so it does not compete with the final correction.

\subsection{Simulation Training and Deployment}
\label{sec:training}
We train with PPO \citep{schulman2017ppo} and generalized advantage estimation
\citep{schulman2016gae} in 4,096 environments. Training rollouts last 12\,s and resample the goal every 4\,s without
resetting the arm, so each rollout chains three attempts and later attempts
start wherever the previous one left the arm.
Appendix~\ref{app:training} gives the architecture and training settings.

\paragraph{Deployment protocol.}
\label{sec:deployment-method}
A strict failure in a single attempt ends either near the goal with a small
residual error or farther from it. When further attempts are allowed, a fixed
protocol handles each case. A failure within 1\,cm of the goal is retried with
an error-corrected goal that cancels the residual. Other failures receive up
to eight attempts that first command a transient goal to move the arm to a
different configuration. Every attempt is judged against the original goal
(Appendix~\ref{app:retry}).

\subsection{Real-Robot Training}
\label{sec:real-implementation}
PolicyIK can also be trained directly on a physical arm, without simulation
weights. The formulation of Sections~\ref{sec:observations}
to~\ref{sec:reward} carries over up to minor hyperparameter changes. On hardware, every
transition costs robot time while computation is relatively cheap, so PolicyIK spends
computation to save transitions and concentrates its data where precision is
decided. We adapt FlashSAC \citep{kim2026flashsac}, an off-policy soft
actor-critic variant \citep{haarnoja2018sac}, together with Hindsight Experience Replay (HER)~\citep{andrychowicz2017hindsight}. Appendix~\ref{app:real_training}
gives all details of this recipe. Below, we highlight some of our design choices.

\paragraph{Latent-alignment critic (Appendix~\ref{app:rt_adapt}).}
Large-rotation goals often require a detour in joint space, during which the pose error first grows.
A standard critic tends to assign similar values to poses that look close, so it favors moving straight
toward the goal. Following LAN~\citep{kang2026lavl}, our critic embeds the state and action with one
network and the goal with another, and predicts the value from their difference (Figure~\ref{fig:rt_arch}b).
Such an architectural design has been shown to better capture the temporal distance between observations and goals~\citep{kang2026lavl}.

\paragraph{Near-goal sampling (Appendix~\ref{app:rt_replay}).}
Steps near the goal decide success but make up a small part of each episode early in training, which is mostly spent on
the approach. We therefore draw most of each training batch from such steps. This focuses learning on
the final correction and holding, and in our experiments it also improved reaching overall.

\paragraph{Reverse curriculum (Appendix~\ref{app:rt_tasks}).}
From random starts, the arm would rarely reach the tolerance early in training. Motivated by
\citet{pmlr-v78-florensa17reverse}, episodes start close to their goals, so the policy learns the final
correction and holding first. As training proceeds, start distances then grow. We also mix this curriculum with sampling from large-rotation tasks directly to encourage the policy to cover wide task distributions.

\vspace{-0.1cm}
\section{Experiments}
\vspace{-0.1cm}
\label{sec:experiments}
We evaluate the precision of PolicyIK in simulation, the contribution of each
component, and its use as a low-level controller in simulated pick-and-place
and on a physical arm.

\vspace{-0.1cm}
\subsection{Simulation Experiments}
\label{sec:simulation}
\label{sec:setup}
\paragraph{Setup and metrics.}
We evaluate a fixed-base 7-DoF OpenArm in Isaac Sim on 4,120 start--goal pairs
from collision-free configurations disjoint from the training pools, with goals
inside the deployment workspace. Each attempt lasts 160 policy steps (5.3\,s).
We report strict and coarse success (Section~\ref{sec:problem}) and median
terminal-window errors over all tasks, including failures, as mean $\pm$
standard deviation across three training seeds. Appendices~\ref{app:mdp}
and~\ref{app:metrics} give the split and aggregation details.

\vspace{-0.1cm}
\paragraph{Baselines.}
Prior RL reaching controllers report position goals or centimeter-level
full-pose tracking (Section~\ref{sec:related}), so we adapt the full-pose
tracking recipe of UMI on Legs~\citep{umi2024} to our fixed-base arm (UMI-RL), retaining its reward, orientation curriculum, network,
and PPO settings with a comparable training budget. Additional tuning is reported in Appendix~\ref{app:umi}.
We also compare against classical differential IK controllers, damped
least squares (DLS) and resolved-rate quadratic programming (QP), and a kinematic
model-predictive controller (MPC), described in Appendix~\ref{app:classical-sim}.

\subsubsection{Main Results}
\label{sec:capability}
\begin{table}[t]
\centering
\caption{Simulation results. With deployment, each task reports its first
strictly successful attempt, or its last attempt if none succeeds, and $E[N]$
is the mean number of attempts. DLS, QP, and MPC are tuned on a separate set
of tasks from the same distribution (Appendix~\ref{app:classical-sim}).}
\label{tab:main}
\small
\setlength{\tabcolsep}{4pt}
\begin{tabularx}{\linewidth}{l*{5}{>{\centering\arraybackslash}X}}
\toprule
 & & & \multicolumn{2}{c}{Median error} & \\
\cmidrule(lr){4-5}
Controller & Strict (\%) $\uparrow$ & Coarse (\%) $\uparrow$ & Pos. (mm) $\downarrow$ & Ori. ($^\circ$) $\downarrow$ & $E[N]$ \\
\midrule
UMI-RL (tuned) & 0.00\sd{0.00} & 16.38\sd{3.10} & 16.65\sd{0.80} & 4.25\sd{0.77} & 1.000 \\
DLS & $50.41$ & $64.42$ & $0.94$ & $0.15$ & 1.000 \\
QP & $56.21$ & $63.98$ & $0.57$ & $0.11$ & 1.000 \\
MPC & $56.33$ & $64.78$ & \underline{$0.54$} & \underline{$0.10$} & 1.000 \\
\rowcolor{policyikrow} PolicyIK & \underline{93.49}\sd{0.20} & \underline{96.56}\sd{0.16} & \textbf{0.25}\sd{0.02} & \textbf{0.04}\sd{0.00} & 1.000 \\
\rowcolor{policyikrow} \quad $+$ deployment & \textbf{98.92}\sd{0.23} & \textbf{99.43}\sd{0.09} & \textbf{0.25}\sd{0.02} & \textbf{0.04}\sd{0.00} & 1.212 \\
\bottomrule
\end{tabularx}
\end{table}

\paragraph{Precision and coverage.}
In a single attempt, PolicyIK achieves 93.5\% strict success and 96.6\% coarse
success, with median position and orientation errors of 0.25\,mm and
$0.04^\circ$, respectively (Table~\ref{tab:main}). The deployment protocol
further increases strict success to 98.9\% through corrected-goal and
transient-goal retries (Appendix~\ref{app:retry}). By comparison, UMI-RL
achieves 16.4\% coarse success after task-specific tuning, but no strict
success. PolicyIK exceeds DLS, QP, and MPC by 43, 37, and 37 points in strict
success. On 34--35\% of tasks, these controllers fail even the coarse criterion
while a joint remains close to its limit during the final second. These cases
account for 71--81\% of their strict failures and typically exhibit nearly
constant terminal errors. PolicyIK achieves about 90\% strict success on the
same tasks (Appendix~\ref{app:classical-sim}).

\vspace{-0.1cm}
\paragraph{Holding.}\label{sec:behavior} PolicyIK stays within both strict thresholds at every sample of the final second
on 92.4\% of tasks, showing that its terminal accuracy extends beyond the
mean-error criterion. On successful tasks, median net motion over that second
is 14\,$\mu$m and $0.002^\circ$ (Appendix~\ref{app:metrics}).

\vspace{-0.1cm}
\paragraph{Efficiency.}
PolicyIK is inexpensive to execute: a forward pass of its two-layer actor takes
0.035\,ms at batch size one with PyTorch on one AMD EPYC 9454 CPU core, three orders of magnitude below the 33\,ms control period, and the
deployment protocol adds
only 0.21 attempts per task on average.

\subsubsection{Ablations}
\label{sec:sources}
\begin{figure}[t]
\centering
\includegraphics[width=\linewidth]{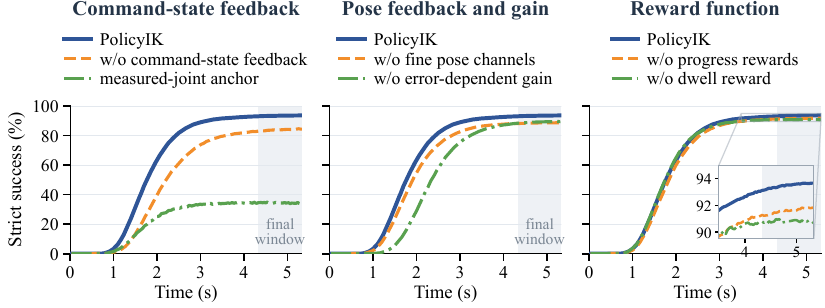}
\vspace{-0.65cm}
\caption{Ablations on command-state feedback, pose feedback and gain, and rewards. Curves show the share of test tasks within 1\,mm and $0.2^\circ$ at each time step.}
\vspace{-1em}
\label{fig:terminal-accuracy}
\end{figure}

\paragraph{Command-state feedback and reference anchor.}
Terminal accuracy is more sensitive to the reference interface than coarse reaching (Figure~\ref{fig:terminal-accuracy}; Table~\ref{tab:ablation}). Removing command-state feedback reduces strict success by 9.2 points. Re-anchoring the reference of this 47-dimensional variant at the measured joints retains 95.9\% coarse success but reduces strict success from 84.3\% to 30.8\%.
These comparisons support retaining the accumulated reference state.

\vspace{-0.1cm}
\paragraph{Pose feedback and action scaling.}
Fine pose channels and error-dependent scaling support complementary aspects
of terminal correction. Removing the fine channels increases median position
error from 0.25 to 0.36\,mm; replacing the error-dependent gain with a fixed gain
increases median entry time from 1.73 to 2.28~s. Either change reduces strict
success by 4--5 points. Appendix~\ref{app:ablation-details} reports
the combined removal and gain-selection analysis.

\vspace{-0.1cm}
\paragraph{Reward function.}
The reward ablations distinguish entering the tolerance from final-window
containment. Without the dwell reward, the arm enters the tolerance about as fast but
leaves it more often before the attempt ends. Among tasks that enter the
tolerance, 7.8\% are not held throughout the final second, against 2.4\%
with dwell.
Removing the progress rewards reduces strict success by 1.8 points.
These results support rewarding both progress and sustained accuracy.

\subsubsection{Pick-and-Place with a VLA Policy}
\label{sec:pick-place}
\paragraph{Setup.}
We evaluate $\pi_{0.5}$-LIBERO~\citep{pi05_2025} on ten adapted
LIBERO-Object tasks~\citep{libero2023} with a side-mounted OpenArm in Isaac Sim.
PolicyIK is retrained for this mount, where gravity acts along a different
base axis, with the same recipe, without pick-and-place data. All controllers use the same fixed model of the Panda operational-space controller to convert the VLA commands into pose references. We compare PolicyIK with default DLS~\citep{wampler1986}, which clamps joint targets, and hand-tuned DLS and QP with joint-limit constraints, on 50 official initial states per task, three runs
each. Appendix~\ref{app:pick-place} details the adapter and controller settings.

\begin{figure}[h]
\centering
\begin{minipage}[t]{0.42\linewidth}
\vspace{0pt}
\centering
\setlength{\abovecaptionskip}{0pt}
\captionof{table}{Pick-and-place success on ten LIBERO-Object tasks in
simulation.}
\label{tab:pick-place}
\vspace{4pt}
\footnotesize
\setlength{\tabcolsep}{4pt}
\renewcommand{\arraystretch}{1.3}
\begin{tabularx}{\linewidth}{>{\raggedright\arraybackslash}Xc}
\toprule
Low-level controller & Success (\%) \\
\midrule
DLS (default) & 90.8 \\
DLS (hand-tuned) & 96.7 \\
QP (hand-tuned) & 90.8 \\
\rowcolor{policyikrow} PolicyIK & 95.7 \\
\bottomrule
\end{tabularx}
\end{minipage}\hfill
\begin{minipage}[t]{0.56\linewidth}
\vspace{0pt}
\centering
\setlength{\abovecaptionskip}{6.5pt}
\includegraphics[width=\linewidth]{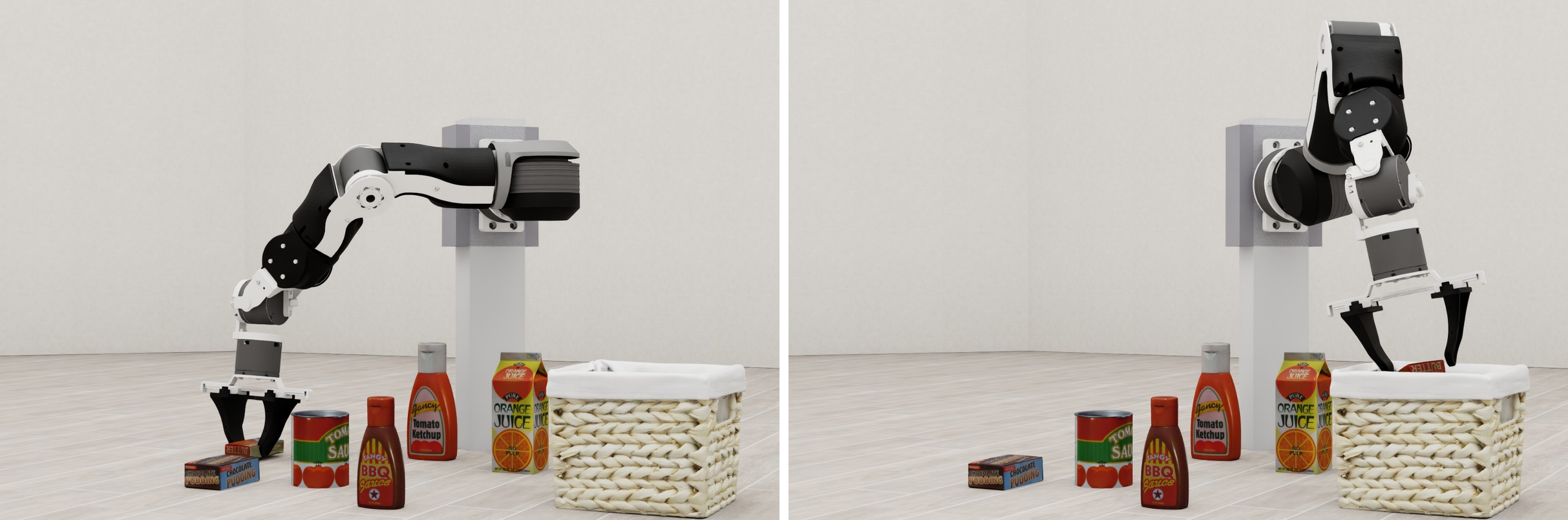}
\captionof{figure}{PolicyIK successfully grasping the butter at its shoulder limit (left) and placing it in the basket (right).}
\label{fig:pick-place}
\end{minipage}
\end{figure}

\vspace{-0.1cm}
\paragraph{Results.}
PolicyIK achieves 95.7\% success (Table~\ref{tab:pick-place}), comparable to hand-tuned DLS (96.7\%) and above default DLS and hand-tuned QP (both 90.8\%). Differences
are concentrated on targets near the mount, where grasping brings a shoulder
joint to its limit (Figure~\ref{fig:pick-place}; Appendix~\ref{app:pick-place}).
Trained only for pose regulation, PolicyIK executes VLA commands in
pick-and-place without a Jacobian, online optimization, or gravity
feedforward.

\vspace{-0.25cm}
\subsection{Real-Robot Experiments}
\label{sec:hardware}

\begin{figure}[ht]
\centering
{\small
\setlength{\tabcolsep}{4pt}
\renewcommand{\arraystretch}{1.15}
\captionof{table}{Reaching, holding, and median pose errors on 60 physical
tasks within 25~s. The best result is in bold, and the second-best is underlined.}
\label{tab:hardware-comparison}
\begin{tabular}{lcccc}
\toprule
Controller & Reached $\uparrow$ & Held $\uparrow$ & Position (mm) $\downarrow$ & Orientation ($^\circ$) $\downarrow$ \\
\midrule
DLS & \underline{48/60} & \underline{17/60} & 0.65 & \underline{0.13} \\
Resolved-rate QP & \underline{48/60} & 16/60 & \underline{0.57} & 0.14 \\
Kinematic MPC & 46/60 & 15/60 & 0.81 & 0.18 \\
Adapted OpenArm QP & 32/60 & 11/60 & 0.99 & 0.23 \\
\midrule
\rowcolor{policyikrow} PolicyIK & \textbf{54/60} & \textbf{43/60} & \textbf{0.50} & \textbf{0.12} \\
\bottomrule
\end{tabular}
\par}
\includegraphics[width=\linewidth]
{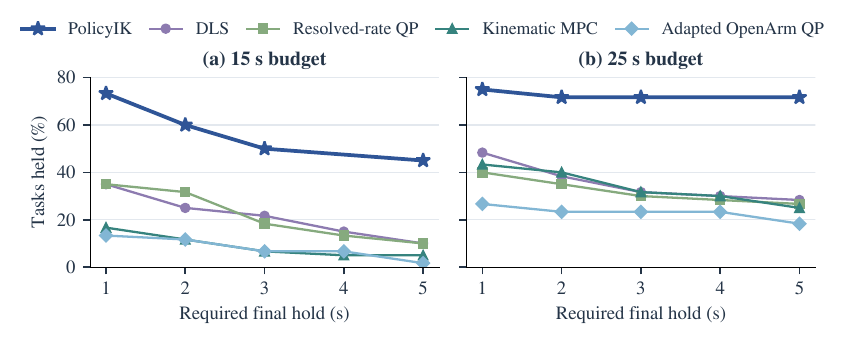}
\vspace{-0.65cm}
\caption{Sustained holding on 60 physical OpenArm tasks as the required final hold increases from 1 to 5 s, at execution budgets of (a) 15 s and (b) 25 s. Both pose errors must remain within 1 mm and $0.2^\circ$ throughout the required interval. Markers denote evaluated holding windows.}
\vspace{-1em}
\label{fig:hardware-hold-sensitivity}
\end{figure}

\begin{wrapfigure}[23]{r}{0.46\textwidth}
\centering
\vspace{-0.6cm} 
\includegraphics[width=0.46\textwidth]{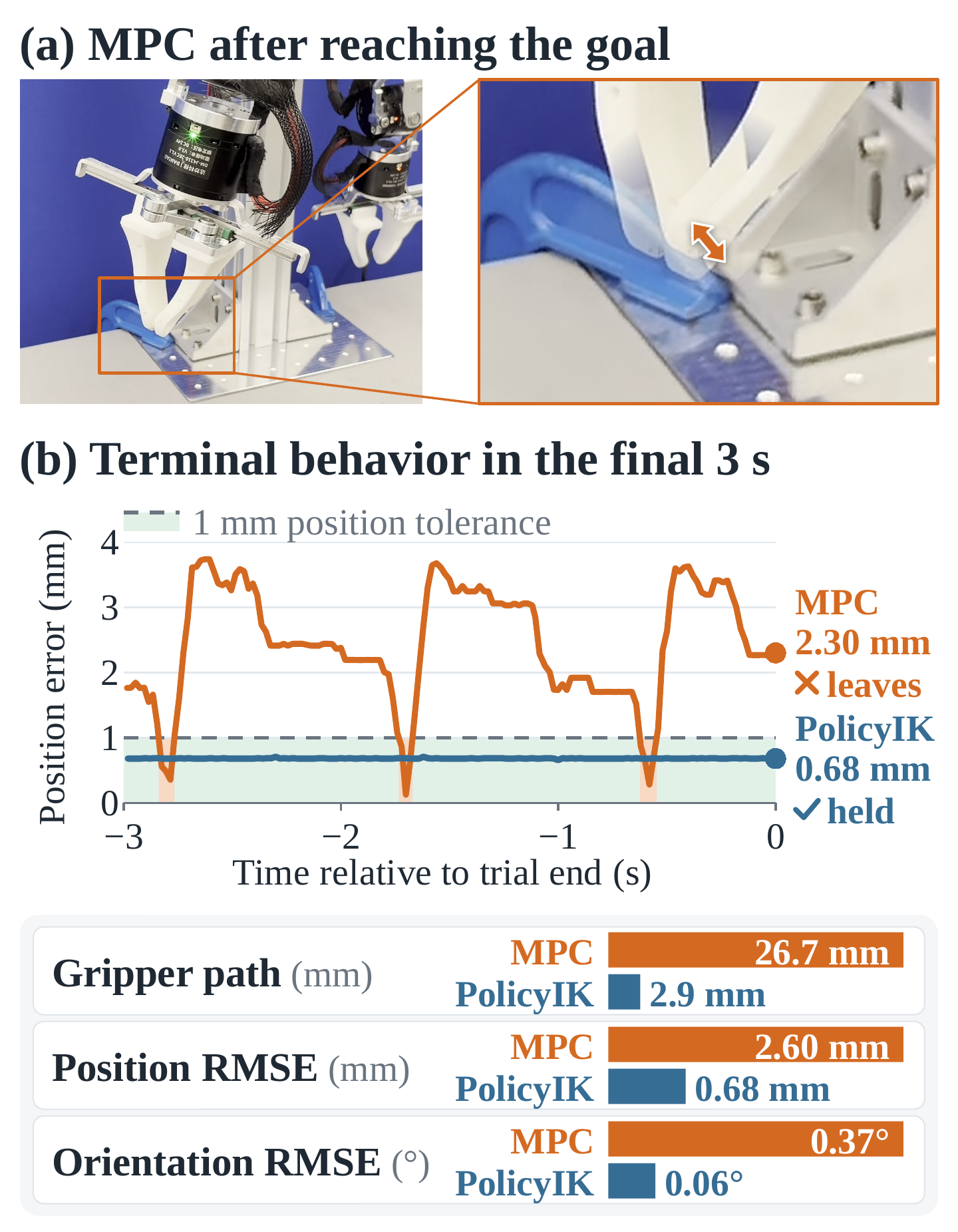}
\caption{A case in which MPC jitters around the goal after reaching, while PolicyIK holds.}
\label{fig:mpc-example}
\end{wrapfigure}

\paragraph{Setup.}
We evaluate PolicyIK trained directly on a physical 7-DoF OpenArm, without sim-to-real transfer (Section~\ref{sec:real-implementation}). We choose OpenArm because its limited wrist motion makes full-pose control challenging~\citep{openarm_v1_specs}. Joint-position references are tracked by PD control with gravity feedforward. Hardware pose errors are computed from joint encoders using forward kinematics in the modeled gripper frame. We mainly evaluate 60 full-pose tasks with a 25~s execution budget.

\paragraph{Metrics.}
A task is reached once both the position and orientation errors remain within 1~mm and $0.2^\circ$ for at least 0.5~s. Holding success requires both errors to remain within these tolerances throughout the final 5~s. Both errors are averaged over the final second of each task and reported as medians across tasks. Appendices~\ref{app:rt_eval} and~\ref{app:baselines} specify the evaluation protocol and controller settings.

\paragraph{Baselines.}
We compare four classical controllers that output joint-position references:
DLS, resolved-rate QP, MPC, and adapted OpenArm QP. Their hardware-tuned
settings and execution constraints are given in Appendix~\ref{app:baselines}.

\vspace{-0.1cm}
\paragraph{Reaching and holding.}
PolicyIK achieves the highest reaching and sustained-holding success among the evaluated controllers. Within 25~s, it reaches 54 of the 60 tasks and holds 43 throughout the final 5~s, achieving 71.7\% holding success compared with at most 28.3\% for the classical controllers (Table~\ref{tab:hardware-comparison}). Extending the required hold from 1~s to 5~s preserves 43 of PolicyIK's 45 successes, showing that it maintains precise control under a longer holding requirement (Figure~\ref{fig:hardware-hold-sensitivity}). Its median position and orientation errors are also the lowest among the evaluated controllers.

Figure~\ref{fig:mpc-example} shows why holding matters after reaching. MPC continues to correct small errors and repeatedly moves outside the 1~mm position tolerance, while PolicyIK stays within. PolicyIK's dwell reward encourages staying within tolerance. Reference accumulation keeps the joint-position command unchanged when actions are zero. The simulation ablations justify both components in holding pose. Appendix~\ref{app:rt_eval} reports the additional execution budgets and holding windows.

\paragraph{Cross-embodiment applicability.}
The PolicyIK formulation also supports precise control on a physical xArm7. Both simulation-trained and hardware-trained policies achieve 60/60 successes on standard reaching and on large-rotation tasks (Appendix~\ref{app:xarm}).

\FloatBarrier
\vspace{-0.05cm}
\section{Conclusion}
\vspace{-0.15cm}
We presented PolicyIK, a goal-conditioned RL formulation for high-precision IK through a joint-position interface. Reference accumulation with command-state feedback retains the offset that the position controller needs at rest, two-scale pose errors with an error-dependent gain resolve millimeter residuals, and progress and dwell rewards train reaching and holding. Ablations show that each design mainly raises terminal precision, so precise IK training should target the terminal phase. In simulation, a single attempt meets the 1\,mm and $0.2^\circ$ tolerance on far more held-out tasks than tuned classical controllers. Training on a physical arm applies the same principle through a reverse curriculum and near-goal sampling. The resulting policy reaches and holds more tasks than all four classical baselines. PolicyIK also executes $\pi_{0.5}$ commands in simulated pick-and-place without task-specific training. These results show that RL can provide high-precision IK for robot arms.

\subsubsection*{Author Contributions}
Yuhan Liu designed the policy's observations, actions, and reward
(Sections~\ref{sec:observations}--\ref{sec:reward}) and conducted the simulation
experiments, including pick-and-place with $\pi_{0.5}$. Yuhan Liu
also led the writing of the paper, organized the codebase, and built the project
website. Xinqi Liu set up the real-robot platform, implemented and tuned the
classical controllers and PolicyIK controller, conducted the OpenArm and xArm7 experiments, drafted the real-robot
sections, and wrote Appendix~\ref{app:baselines}. Zixuan Dong extended the
training recipe to hardware with FlashSAC, conducted the OpenArm experiments,
wrote Appendix~\ref{app:real_training}, and revised
Sections~\ref{sec:problem} and~\ref{sec:real-implementation}.

\bibliographystyle{preprint}
\bibliography{references}
\clearpage
\appendix
\etocdepthtag.toc{appendix}
\etocsettagdepth{mainbody}{none}
\etocsettagdepth{appendix}{subsection}
\etocsettocstyle{\section*{Appendix Contents}}{}
\tableofcontents
\clearpage

\section{Simulation Setup}
\label{app:mdp}
\paragraph{Controller, observation, and action.}
Physics runs at 60~Hz and the policy at 30~Hz. Between policy steps, an implicit PD controller with
stiffness 80 and damping 4 on every arm joint tracks the reference under gravity
and finite effort limits, without feedforward compensation, and the solver uses
32 position iterations. This follows the joint-position reach task of Isaac Lab.
With gravity feedforward added to the PD targets, PolicyIK reaches 94.3\% strict
success in a single attempt, against 93.5\% without it. The reference starts at the measured joint positions. The
action scale is $\eta=0.0625$. In Equation~\ref{eq:action-gain},
$\beta_{\min}=0.1$, $(l_p,u_p)=(0.5,9.31)$~mm, and
$(l_R,u_R)=(0.415,2.328)^\circ$, with angular computations in radians.

The keypoint offsets in Equation~\ref{eq:keypoints} are 0.3~m along each
end-effector axis, and the observation in Equation~\ref{eq:observation} has
54 dimensions. During training, uniform noise in $[-0.01,0.01]$ perturbs the
joint-position and joint-velocity channels. Evaluation uses noise-free
observations.

\paragraph{Holding offset.}
For an unsaturated PD controller without feedforward, static balance gives
\begin{equation}
K_p(z^*-q^*)=\tau_g(q^*),\qquad
z^*=q^*+K_p^{-1}\tau_g(q^*).
\label{eq:reference-equilibrium}
\end{equation}

\paragraph{Data splits.}
\label{app:data}
The source set contains 200,000 joint configurations sampled uniformly within
the joint limits, together with their forward-kinematics poses. After
floor-contact and self-collision filtering and the removal of all evaluation
rows, training starts are sampled uniformly over 157,504 rows and goals over the
104,599 rows whose end effector lies in the deployment workspace. This workspace
is a cylinder about the base axis with a radius of 0.49~m and heights from 0.15
to 0.50~m, excluding a keep-out region around the base links within 0.128~m of
the axis and below 0.222~m. Evaluation uses 4,120 fixed start--goal pairs
sampled without replacement under the same rules. Training and evaluation share
no configuration, and no evaluation goal lies within both 1~mm and $0.2^\circ$
of a training goal.

\section{Reward and Simulation Training}
\label{app:training}
\paragraph{PPO and checkpoint averaging.}
Each PPO epoch collects 24 steps from each of 4,096 environments and makes 5
passes over these 98,304 transitions in minibatches of 24,576. Training runs
13,000 epochs, about $1.28\times10^9$ transitions. The actor and critic share a
two-layer 256-unit ELU network, and the policy learns a state-independent log
standard deviation. PPO uses $\gamma=0.99$, GAE $\lambda=0.95$, clip ratio 0.2,
entropy coefficient 0.01, value-loss coefficient 2, and an adaptive learning
rate initialized at $10^{-3}$ with KL target 0.01. The deployed policy is an
exponentially weighted average of the last 20 checkpoints, saved every 100
epochs, with weight proportional to $0.5^{\ell/400}$ for the checkpoint $\ell$
epochs before the last.

\paragraph{Error history.}
The reward at state $t$ corresponds to action $a_{t-1}$ and uses the previous
best $b_{t-1}$ before the update $b_t=\min(b_{t-1},\rho_t)$. The observation
exposes the same quantities through
\begin{equation}
h_t=\left[\operatorname{clip}(b_{t-1}/20,0,1),\;
\operatorname{clip}((\rho_t-b_{t-1})/20,0,1)\right].
\label{eq:history}
\end{equation}
On each new goal, the running best is initialized to the current $\rho$, and the
improvement, backslide, potential-shaping, and dwell terms are zeroed for that
step.

\paragraph{Reward parameters.}
Equation~\ref{eq:total-reward} groups the eleven implemented terms. The keypoint offsets of the tracking reward in Equation~\ref{eq:tracking-reward}
are $\tilde c_i=\kappa(e_i-\frac13\mathbf{1})$ with $\kappa=0.45\,\mathrm{m}$.
The regularization costs are
\begin{equation}
\begin{split}
c_t^{\mathrm{motion}}&=w_a(t)\|a_{t-1}-a_{t-2}\|_2^2+w_v(t)\|\dot q_t\|_2^2,\\
c_t^{\mathrm{limit}}&=w_m\operatorname{clip}
\left(1-\frac{\min_j m_j(q_t)}{0.05},0,1\right)\mathds{1}[\rho_t>10],\\
m_j(q)&=\frac{\min(q_j-q_j^{\mathrm{lo}},q_j^{\mathrm{hi}}-q_j)}
{q_j^{\mathrm{hi}}-q_j^{\mathrm{lo}}}.
\end{split}
\end{equation}
The motion-cost weights follow the Isaac Lab reward-weight schedule:
$(w_a,w_v)=(5\times10^{-4},5\times10^{-4})$ until the environment step counter
exceeds 4,500 policy steps, and $(0.025,0.005)$ afterwards. The counter
restarts in each of the three training segments (epochs 0 to 3,500, 3,500 to
8,500, and 8,500 to 13,000), so the larger weights apply for about 95\% of
training.
The progress weights are $w_n=26.831$ and $w_b=0.2$, and the joint-margin
weight is $w_m=1$.
The reward manager multiplies every term by the policy period of 1/30~s,
which scales all weights equally.

\section{Metrics and Deployment Protocol}
\label{app:metrics}
\paragraph{Metrics.}
Beyond the success criteria of Section~\ref{sec:problem}, three measures
describe the terminal phase. A task \emph{enters} the strict region when one
sample meets both strict thresholds, and its entry time is the time of the
first such sample. A task is \emph{contained} when every sample of the final
window meets both thresholds. Among the tasks that enter, we report the share
that is not contained. The final-second net motion is the distance between the end-effector positions
at the start and end of the final second, together with the rotation angle
between the corresponding orientations.

Each statistic is computed separately for each training seed. Across seeds,
we report its mean and sample standard deviation.

\paragraph{Deployment protocol.}
\label{app:retry}
Each attempt is a 160-step episode from the task's start configuration. When a
failed first attempt ends with a terminal-window position error within 1~cm,
the error-corrected goal shifts the requested position by the negative
final-step position error of that attempt and rotates the requested orientation
by the inverse of its final-step rotation error. A transient goal displaces the
requested position by 0.30~m along a random unit direction and rotates the
requested orientation by $90^\circ$ about a random unit axis. It is commanded for
the first 40 steps, and the original goal for the remaining 120 steps. Each of
the eight transient-goal attempts draws its directions and axes per task from
its own fixed seed.

Table~\ref{tab:deployment} separates the contributions of checkpoint averaging
and the two protocol steps. Per seed, a single attempt leaves on average 127
tasks that meet only the coarse criterion and 142 that miss it. The
error-corrected goal recovers 54 tasks, nearly all from the first group, and
transient goals recover 170 more, including 117 from the second. The protocol
thus acts on the tail of the error distribution: it lowers the mean position error by a factor of five, while the median stays at 0.25~mm. The
45 tasks per seed that still fail have goal configurations closer to joint
limits: the worst joint of the goal configuration lies a median of 1.3\% of its
range from a limit, against 5.7\% for tasks solved in the first attempt.

\begin{table}[t]
\centering
\caption{Strict success as checkpoint averaging and the two deployment steps are
added. $E[N]$ is the mean number of attempts per task.}
\label{tab:deployment}
\small
\begin{tabular}{lcc}
\toprule
Setting & Strict (\%) $\uparrow$ & $E[N]$ \\
\midrule
Final checkpoint, single attempt & 92.30\sd{0.13} & 1.000 \\
Checkpoint average, single attempt & 93.49\sd{0.20} & 1.000 \\
$+$ error-corrected goal & 94.80\sd{0.12} & 1.036 \\
$+$ transient-goal attempts & \textbf{98.92}\sd{0.23} & 1.212 \\
\bottomrule
\end{tabular}
\end{table}

\section{Simulation Baseline}
\label{app:umi}
UMI-RL converts the per-step constants of the source recipe from its 50~Hz
policy rate to our 30~Hz, so that the discount horizon and rollout span stay
fixed in seconds. It trains for 20,000 epochs of 14 steps in 4,096
environments, $1.15\times10^9$ transitions against $1.28\times10^9$ for
PolicyIK.

\paragraph{Tuning.}
Table~\ref{tab:umi_kernel} summarizes the task-adapted settings. Its last row, a discount-horizon-matched configuration with $\gamma=0.99$, $\lambda=0.95$, and $k=0.1$, did not improve strict or coarse success over the best tuned UMI-RL variant.

\begin{table}[ht]
\centering
\caption{Tuned variants of UMI-RL. Entries are means and standard deviations
across three training seeds. Position and orientation errors are median
terminal-window errors. The factor $k$ scales the orientation ladder of the source recipe.
Strict success is zero in every setting.}
\label{tab:umi_kernel}
\small
\setlength{\tabcolsep}{4pt}
\begin{tabular}{lccc}
\toprule
 & & \multicolumn{2}{c}{Median error $\downarrow$} \\
\cmidrule(lr){3-4}
Setting & Coarse (\%) $\uparrow$ & Pos. (mm) & Ori. ($^\circ$) \\
\midrule
0.20~s horizon, $k=1$ (source) & 0.04\sd{0.01} & 19.45\sd{3.80} & 74.69\sd{2.62} \\
0.54~s horizon, $k=1$ & 0.02\sd{0.02} & 6.27\sd{0.43} & 54.87\sd{3.85} \\
0.54~s horizon, $k=0.5$ & 0.35\sd{0.09} & 11.17\sd{2.27} & 37.45\sd{0.61} \\
0.54~s horizon, $k=0.25$ & 16.38\sd{3.10} & 16.65\sd{0.80} & 4.25\sd{0.77} \\
3.3~s horizon, $k=0.1$ & 1.04\sd{0.06} & 44.16\sd{3.52} & 4.20\sd{1.25} \\
\bottomrule
\end{tabular}
\end{table}

\paragraph{Deployment protocol.}
Applied to the source-horizon configuration, the deployment protocol achieves
$0.02\pm0.01$\% strict and $0.05\pm0.02$\% coarse success, with 9.196 attempts
per task on average.

\section{Classical Controllers in Simulation}
\label{app:classical-sim}
The simulated DLS, QP, and MPC follow the control laws, task weighting $W$, and
damping $\lambda$ of the classical controllers in Appendix~\ref{app:baselines}.
The MPC keeps the seven-step horizon and objective of the hardware MPC and is
solved for each arm with DAQP.
Each integrates the next joint reference from the previous one under
joint-position and command-lead bounds. All run at the 30~Hz policy rate and
receive exact gravity compensation from the simulator, which PolicyIK does not
receive. The reference-speed bound used for safety on the physical arm is not
applied, so the classical controllers can use the full 5.3~s attempt.

\paragraph{Parameter selection.}
The gain $k$ and the command-lead bound of each controller were tuned on a
separate set of tasks drawn from the same distribution as the test tasks and
disjoint from them. The selected values are $k=2$ for DLS and $k=1.5$ for QP and
MPC, all with a 0.15~rad lead bound, and the MPC acceleration bound is raised to
100~rad/s$^2$. Table~\ref{tab:classical-sim} lists the selected settings and
further variants, and adding a
null-space joint-limit objective did not improve strict success.

\begin{table}[h]
\centering
\caption{Classical controllers on the 4,120 test tasks. The selected settings
appear in Table~\ref{tab:main}. DLS (default) is the Isaac Lab
differential IK controller with its default settings, as in Section~\ref{sec:pick-place}.}
\label{tab:classical-sim}
\small
\setlength{\tabcolsep}{4pt}
\begin{tabular}{lcccc}
\toprule
 & & & \multicolumn{2}{c}{Median error $\downarrow$} \\
\cmidrule(lr){4-5}
Controller & Strict (\%) $\uparrow$ & Coarse (\%) $\uparrow$ & Pos. (mm) & Ori. ($^\circ$) \\
\midrule
DLS, selected ($k=2$, 0.15~rad lead) & 50.41 & 64.42 & 0.94 & 0.15 \\
QP, selected ($k=1.5$, 0.15~rad lead) & 56.21 & 63.98 & 0.57 & 0.11 \\
MPC, selected ($k=1.5$, 0.15~rad lead) & 56.33 & 64.78 & 0.54 & 0.10 \\
DLS, measured-joint anchor, full step & 1.84 & 46.50 & 4.74 & 2.20 \\
DLS (default), 30~Hz & 1.60 & 42.86 & 5.88 & 2.73 \\
DLS (default), 60~Hz & 1.63 & 43.18 & 5.79 & 2.71 \\
DLS (default), 100~Hz & 4.95 & 54.47 & 3.16 & 1.52 \\
\bottomrule
\end{tabular}
\end{table}

\begin{figure}[h]
\centering
\includegraphics[width=\linewidth]{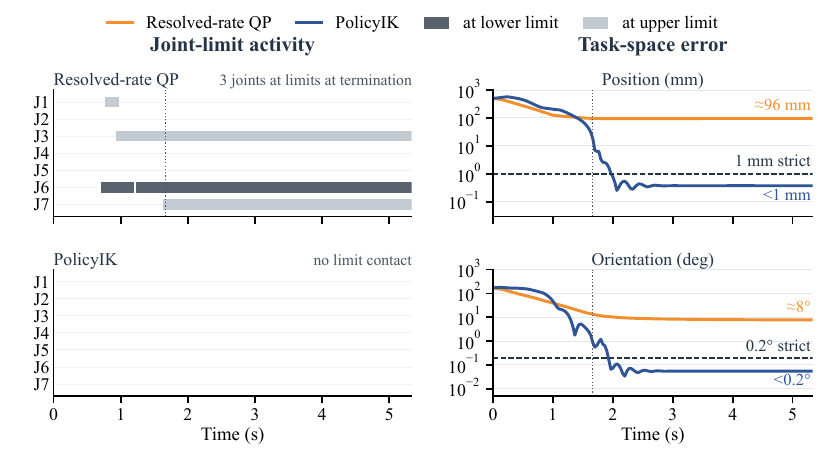}
\caption{One test task. Left: intervals in which each joint lies within
0.01~rad of its lower (dark) or upper (light) limit. Right: position and
orientation errors, with the strict tolerances dashed. The dotted line marks
when a third QP joint reaches its limit (1.7~s). The QP failure coincides with
this sustained limit occupancy, and its error then stays near 96~mm and
$8^\circ$. PolicyIK never reaches a joint limit and enters the strict
tolerance at 2.0~s.}
\label{fig:classical-limit}
\end{figure}

\paragraph{Failure analysis.}
We identify the joint-limit stalls discussed in Section~\ref{sec:capability}
by coarse failure and a joint remaining within 0.01~rad of a soft limit for
at least half of the final second.
In these failures, joint constraints restrict the local updates, and the pose
error plateaus before reaching the goal.
Figure~\ref{fig:classical-limit} illustrates this contrast.

\section{Ablation Details}
\label{app:ablation-details}
Table~\ref{tab:ablation} lists the full results of the component ablations in
Section~\ref{sec:sources}. Removing command-state feedback deletes its seven
channels from the 54-dimensional observation. The measured-joint anchor uses
the same 47 dimensions and replaces the anchor $z_{t-1}$ in
Equation~\ref{eq:reference-update} with $q_t$, so the comparison between the
two isolates the anchor. Removing the fine pose channels leaves 48 dimensions.
Removing channels shrinks the input layer, while hidden widths and training
budgets stay fixed. Removing the progress group disables the improvement,
potential-shaping, and backslide rewards but keeps the running-best
observation, and removing dwell sets only its reward weight to zero.

\begin{table}[t]
\centering
\caption{Component ablations with one attempt per task and matched training
budgets. Entry is the median first-entry time and Uncont. the uncontained share
(Appendix~\ref{app:metrics}). The measured-joint anchor sets $z_{t-1}=q_t$ in
Equation~\ref{eq:reference-update} and removes command-state feedback, and the
fixed gain is \mbox{$\beta=0.1$}.}
\label{tab:ablation}
\footnotesize
\setlength{\tabcolsep}{1.5pt}
\begin{tabularx}{\linewidth}{l>{\hsize=1.04\hsize\centering\arraybackslash}X>{\hsize=1.04\hsize\centering\arraybackslash}X>{\hsize=.96\hsize\centering\arraybackslash}X>{\hsize=1.04\hsize\centering\arraybackslash}X>{\hsize=.96\hsize\centering\arraybackslash}X>{\hsize=.96\hsize\centering\arraybackslash}X}
\toprule
 & \multicolumn{2}{c}{Success (\%) $\uparrow$} & \multicolumn{2}{c}{Terminal phase $\downarrow$} & \multicolumn{2}{c}{Median error $\downarrow$} \\
\cmidrule(lr){2-3}\cmidrule(lr){4-5}\cmidrule(lr){6-7}
Variant & Coarse & Strict & Entry (s) & Uncont. (\%) & Pos. (mm) & Ori. ($^\circ$) \\
\midrule
\rowcolor{policyikrow} PolicyIK & 96.56\sd{0.16} & 93.49\sd{0.20} & 1.73\sd{0.01} & 2.43\sd{0.23} & 0.25\sd{0.02} & 0.04\sd{0.00} \\
\midrule
\multicolumn{7}{l}{\textit{Command-state feedback and reference anchor}} \\
w/o command-state feedback & 92.63\sd{0.20} & 84.26\sd{0.28} & 2.04\sd{0.02} & 10.32\sd{0.32} & 0.30\sd{0.00} & 0.06\sd{0.00} \\
measured-joint anchor & 95.90\sd{0.17} & 30.77\sd{7.94} & 1.80\sd{0.03} & 82.46\sd{6.62} & 1.27\sd{0.05} & 0.16\sd{0.01} \\
\addlinespace
\multicolumn{7}{l}{\textit{Pose feedback and action scaling}} \\
w/o fine pose channels & 95.39\sd{0.25} & 88.41\sd{0.34} & 1.83\sd{0.03} & 4.38\sd{0.11} & 0.36\sd{0.01} & 0.06\sd{0.00} \\
w/o error-dependent gain & 94.89\sd{0.14} & 89.13\sd{1.64} & 2.28\sd{0.07} & 5.69\sd{2.62} & 0.26\sd{0.03} & 0.05\sd{0.01} \\
w/o both ($\beta=0.25$) & 93.01\sd{0.59} & 85.27\sd{0.39} & 2.11\sd{0.02} & 4.85\sd{0.33} & 0.34\sd{0.00} & 0.06\sd{0.00} \\
\addlinespace
\multicolumn{7}{l}{\textit{Reward function}} \\
w/o progress rewards & 96.47\sd{0.49} & 91.66\sd{0.71} & 1.79\sd{0.05} & 3.41\sd{0.65} & 0.26\sd{0.00} & 0.04\sd{0.00} \\
w/o dwell reward & 96.87\sd{0.29} & 91.54\sd{1.33} & 1.69\sd{0.02} & 7.83\sd{1.18} & 0.34\sd{0.00} & 0.07\sd{0.00} \\
\bottomrule
\end{tabularx}
\end{table}

\paragraph{Fixed-gain selection.}
Fixed gains are selected on 2,084 development tasks with one seed per
candidate. With fine channels, gains of 0.1, 0.25, 0.5, and 1.0 reach 87.24\%,
85.70\%, 82.63\%, and 82.77\% strict success. Without them, the same gains reach
0.00\%, 79.41\%, 76.20\%, and 74.90\%. Each fixed-gain variant therefore uses its
best gain, 0.1 with fine channels and 0.25 without. At $\beta=0.1$, the variant
without fine channels never enters the strict region, and its median errors are
11.57~mm and $5.55^\circ$.

\section{Pick-and-Place Evaluation Details}
\label{app:pick-place}

\paragraph{Tasks and protocol.}
Each LIBERO-Object task asks the robot to pick one grocery item and place it in
a basket. We use each task's 50 official initial states and language
instruction, the agent-view and wrist cameras at LIBERO's poses, a budget of
280 actions, and the evaluation settings of $\pi_{0.5}$-LIBERO: action chunks
are re-planned every five actions and images are resized to
$224\times224$. Objects start from the official states after LIBERO's settling
interval. An episode succeeds the first time LIBERO's success predicate holds
at an action boundary, and it fails if the fingers pass into the table. Each
controller runs three times per initial state. $\pi_{0.5}$ samples its own
noise in every run, and the three runs of PolicyIK use its three training
seeds.

\paragraph{Pose reference.}
$\pi_{0.5}$ outputs relative end-effector commands for robosuite's
operational-space controller~\citep{robosuite2020} at 20~Hz. A model of that
controller with robosuite's gains (stiffness 150, critical damping) integrates
them into an end-effector pose that is re-anchored at every action, and the
low-level controller tracks this pose at 30~Hz. No parameter of the model is
fitted. The commanded tool-center point is kept at least 5~mm above the table,
and the gripper opens or closes with the sign of the gripper command.

\paragraph{Robot and scene.}
One planar translation and yaw of the robot and its table, chosen from
source-robot trajectories, places the objects within reach, and the arm starts
at the Panda's initial end-effector pose. The cameras render the LIBERO scene
appearance, object collision geometry, mass and inertia follow the LIBERO
assets, and the physics runs at 60~Hz.

\paragraph{Controllers.}
PolicyIK is trained from random initialization for the side-mounted arm of
Section~\ref{sec:pick-place}, with three seeds and checkpoint averaging. Its
goals are sampled from reachable configurations of this arm. The default DLS is the differential inverse-kinematics controller of Isaac Lab~\citep{orbit2023} with damping 0.01, whose joint targets are clamped at the soft limits. The hand-tuned DLS and QP use the hardware settings of Appendix~\ref{app:baselines} (task weighting $W$, damping 0.02 and 0.005, joint
limits as bounds of a bounded least-squares step) and take one full step per
control cycle without the hardware speed and command-lead bounds. All three
classical controllers add gravity-compensation torques and share the arm's
joint drives.

\paragraph{Behavior near joint limits.}
In the evaluated trajectories, the default DLS discards the bound-violating part of
its step and stops short of the commanded pose, while the lightly damped QP
can stall at a joint limit during lifting. PolicyIK completes these grasps
through a reference clipped at the soft limits.

\paragraph{Results by target location.}
Table~\ref{tab:pick-place-tasks} groups the ten tasks by where their targets lie.
The targets of alphabet soup, butter, chocolate pudding, ketchup, and milk lie
closest to the mount, 0.28~m in front of the shoulder and 0.28--0.34~m below it,
where the grasp holds a shoulder joint at its limit. The other five lie a
further 0.21~m to the side.

\begin{table}[h]
\centering
\caption{Pick-and-place success (\%) by target location, three runs from each of
the 50 official initial states per task (750 episodes per group).}
\label{tab:pick-place-tasks}
\footnotesize
\setlength{\tabcolsep}{4pt}
\begin{tabular}{lcccc}
\toprule
Targets & \begin{tabular}[b]{@{}c@{}}DLS\\(default)\end{tabular} & \begin{tabular}[b]{@{}c@{}}DLS\\(hand-tuned)\end{tabular} & \begin{tabular}[b]{@{}c@{}}QP\\(hand-tuned)\end{tabular} & PolicyIK \\
\midrule
Closest to the mount & 85.5 & 96.8 & 85.9 & 95.1 \\
Farther to the side & 96.1 & 96.5 & 95.7 & 96.4 \\
\bottomrule
\end{tabular}
\end{table}

\section{Training PolicyIK on Physical Arms}
\label{app:real_training}

This appendix describes how we train PolicyIK directly on a physical arm with FlashSAC~\citep{kim2026flashsac}.
Appendix~\ref{app:rt_flashsac} reviews FlashSAC, and Appendices~\ref{app:rt_adapt}--\ref{app:rt_env} describe how we
adapt it to a single physical arm: the learner and its critic, replay, training tasks, and environment settings.
Appendix~\ref{app:rt_ablation} ablates the three components that matter on hardware, and
Appendix~\ref{app:rt_eval} gives the evaluation protocol. Table~\ref{tab:rt_params} lists every setting. The OpenArm
and xArm7 (Appendix~\ref{app:xarm}) hardware policies use the same settings.

\subsection{FlashSAC Recap}
\label{app:rt_flashsac}

FlashSAC is an off-policy actor--critic method built on soft actor--critic (SAC)~\citep{haarnoja2018sac}. It was
designed for fast and stable training with many parallel simulators and changes SAC mainly in its network
architecture and its critic objective. We summarize the parts that PolicyIK relies on and refer to
\citet{kim2026flashsac} for the full method. We write $o$ for the observation of Equation~\ref{eq:observation}, which contains the goal,
$\pi_\theta$ for the policy, and $Q_{\omega_1}, Q_{\omega_2}$ for the two critics, with target parameters
$\bar\omega_1, \bar\omega_2$ updated by Polyak averaging at rate $\tau$.

\begin{figure}[ht]
\centering
\includegraphics[width=\linewidth]{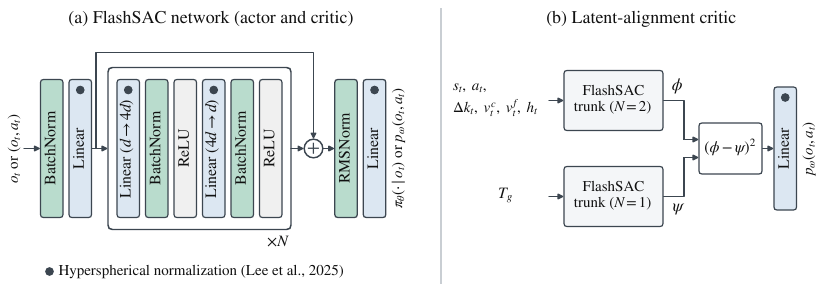}
\caption{(a) The FlashSAC network, used for the actor and, in FlashSAC, for each critic. Filled circles mark the hyperspherical normalization~\citep{lee2025simbav2}, where weight vectors project onto the unit
sphere after every gradient step. (b) The latent-alignment critic that PolicyIK uses on hardware
(Appendix~\ref{app:rt_adapt}). In both panels, $p_\omega(o_t, a_t)$ denotes the value-distribution logits that the
critic's linear head outputs; a softmax over the atoms turns them into the probabilities $p_{\omega, i}$ of
Equation~\ref{eq:rt_critic}.}
\label{fig:rt_arch}
\end{figure}

\paragraph{Architecture.}
Actor and critics share the architecture of Figure~\ref{fig:rt_arch}(a). The input is batch-normalized and linearly
embedded into width $d$. $N$ inverted-bottleneck residual blocks follow, each expanding to width $4d$ and projecting
back to $d$, with batch normalization before each ReLU. RMS normalization bounds the feature norm before a linear
head, which outputs the mean and standard deviation of a tanh-Gaussian policy or the atom logits of a critic. Batch
normalization is preferred over layer normalization because it uses the statistics of large and diverse replay
batches. FlashSAC also applies hyperspherical normalization~\citep{lee2025simbav2}. After every gradient
step, each weight vector of a linear layer is projected onto the unit sphere, and the scale and shift vectors of
each normalization layer are projected to norm $\sqrt{d}$. The network then encodes information in directions rather
than magnitudes, which prevents the weight growth that amplifies estimation errors under bootstrapping.

\paragraph{Critic objective.}
Each critic predicts a categorical distribution over $n_{\mathrm{atom}}$ atoms
$v_i = G_{\min} + (i-1)\Delta$ with $\Delta = (G_{\max} - G_{\min})/(n_{\mathrm{atom}} - 1)$, and its action value is the
mean, $Q_\omega(o, a) = \sum_i p_{\omega, i}(o, a)\, v_i$~\citep{bellemare2017c51}. For a batch of transitions
$(o, a, r, o', d)$ from the replay buffer $\mathcal{D}$, where $d$ marks termination, next actions are drawn as $a' \sim \pi_\theta(\cdot \mid o')$. Batch
normalization makes a network's output depend on the batch it is evaluated with, so FlashSAC evaluates the current and next pairs in one joint forward pass,
\begin{equation}
\left[ \begin{array}{c} p_{\omega}(o, a) \\ p_{\omega}(o', a') \end{array} \right]
= P_{\omega}\!\left( \left[ \begin{array}{c} o \\ o' \end{array} \right],
\left[ \begin{array}{c} a \\ a' \end{array} \right] \right),
\label{eq:rt_joint}
\end{equation}
so both halves share the same normalization statistics. Following clipped double Q
learning, the target uses the distribution $\bar p$ of the target critic with the smaller expected value at
$(o', a')$. Each atom is moved by the soft Bellman backup and clipped to the support,
$\hat v_i = \mathrm{clip}\big( r + \gamma (1 - d) ( v_i - \alpha \log \pi_\theta(a' \mid o') ),\, G_{\min},\, G_{\max} \big)$,
and its probability is split between the two nearest atoms. Each critic $j \in \{1, 2\}$ minimizes the
cross-entropy to this projected target,
\begin{equation}
\mathcal{L}_Q(\omega_j) = -\,\mathbb{E}_{\mathcal{D}}\Big[ \sum_{k=1}^{n_{\mathrm{atom}}} m_k \log p_{\omega_j, k}(o, a) \Big],
\qquad
m_k = \sum_{i=1}^{n_{\mathrm{atom}}} \bar p_i \max\Big( 0,\ 1 - \frac{| \hat v_i - v_k |}{\Delta} \Big).
\label{eq:rt_critic}
\end{equation}

\paragraph{Policy objective.}
The actor maximizes the smaller critic value with an entropy bonus weighted by the temperature $\alpha$,
\begin{equation}
\mathcal{L}_\pi(\theta) = \mathbb{E}_{o \sim \mathcal{D},\, a \sim \pi_\theta(\cdot \mid o)}
\Big[ \alpha \log \pi_\theta(a \mid o) - \min_{j \in \{1, 2\}} Q_{\omega_j}(o, a) \Big].
\label{eq:rt_actor}
\end{equation}
The temperature is tuned so that the policy entropy tracks a target
$\bar{\mathcal{H}}$, which FlashSAC sets to the entropy of a Gaussian with standard deviation $\sigma_{\mathrm{tgt}}$
in each of the $|\mathcal{A}|$ action dimensions,
\begin{equation}
\mathcal{L}(\alpha) = \mathbb{E}_{o \sim \mathcal{D},\, a \sim \pi_\theta(\cdot \mid o)}
\Big[ -\alpha \big( \log \pi_\theta(a \mid o) + \bar{\mathcal{H}} \big) \Big],
\qquad
\bar{\mathcal{H}} = \frac{|\mathcal{A}|}{2} \log\big( 2 \pi e\, \sigma_{\mathrm{tgt}}^2 \big).
\label{eq:rt_temp}
\end{equation}

\subsection{Adapting FlashSAC to a Physical Arm}
\label{app:rt_adapt}

\paragraph{Hyperparameters.}
Table~\ref{tab:rt_params} lists our settings next to FlashSAC's configuration. FlashSAC
targets settings in which data are cheap. On a physical arm, transitions cost robot time and computation is
comparatively cheap, so we spend computation to save transitions. The replay buffer keeps every transition of the
run, updates start as soon as the buffer holds one batch, and we perform more updates per transition. We
use a batch of 1{,}024, half of FlashSAC's batch for parallel simulation, so that two updates fit within one control
period. With more updates per transition, we halve the target update rate to $\tau = 0.005$, which keeps the
bootstrap target steadier. We double the
number of atoms to 201, which halves the atom spacing and resolves the small return differences between holding the
tolerance and drifting just outside it. A single arm also lacks the exploration of many parallel environments, so we
raise the entropy target from $\sigma_{\mathrm{tgt}} = 0.15$ to 0.30. This matches the intuition, and in our
experiments the higher target also avoided a late decline in success during training.

\paragraph{Latent-alignment critic.}
A dense reward on the pose error, and a critic that generalizes across inputs that are close in Euclidean distance,
both favor actions that shrink the end-effector error along a straight path. Large-rotation tasks often require a
detour through joint space instead, along which the error first grows. Their temporal distance to the goal, the
number of steps still needed, then differs from the Euclidean pose distance. This is also why some shaping term of
the reward is potential-based. 

\citet{kang2026lavl} report the same mismatch for
goal-conditioned value functions. An MLP value function assigns similar values to states that are close in Euclidean
distance and therefore propagates values between states that are far apart in temporal distance. Their latent
alignment network (LAN) instead parameterizes the value as a distance between learned representations,
$V(s, g) = -\| \phi_S(s) - \phi_G(g) \|_2$, so that values generalize through the learned latent space. We adapt
this structure to FlashSAC's categorical action-value critic (Figure~\ref{fig:rt_arch}b). A state--action encoder
$\phi$ receives the state $s_t$, the action $a_t$, and the goal-relative channels of $o_t$: the position error, the
keypoint offsets $\Delta k_t$, the coarse and fine errors $v^c_t, v^f_t$, and the running-best features $h_t$. A goal
encoder $\psi$ receives the goal pose $T_g$. Each encoder is a FlashSAC trunk, the network of
Figure~\ref{fig:rt_arch}(a) without its output layer, of width 256 with two and one residual blocks, respectively.
The categorical head receives the element-wise squared difference $(\phi - \psi)^2$ in place of the single trunk's
features. The goal-relative channels depend on both the state and the goal, and they
must enter the state--action encoder.
Without them there, the action and the goal would meet only in the final linear layer.

One update of the actor and the latent-alignment critic takes about 27\,ms on the onboard GPU of our controller
(NVIDIA Jetson Thor). Two updates, including batch assembly, therefore fit within the 100\,ms control period, and the
learner runs them while the arm tracks the reference it has just received.

\paragraph{Update to data ratio (UTD).}
More gradient updates per transition generally improve sample efficiency, which matters most when every transition
costs robot time, but each update costs wall-clock time. We use two updates per control step by default. Between
episodes, the arm moves to the next start configuration and settles, which takes 5.5\,s, and the learner fills this
time with as many additional updates as fit, about 81 per reset. A 150-step episode thus receives about 381 updates,
or 2.5 per transition, at no cost in robot time.

\subsection{Replay}
\label{app:rt_replay}

\paragraph{Batch composition.}
Each batch draws 25\% of its transitions from a fixed demonstration buffer and 75\% from online data, the
fixed-ratio mixing of RLPD~\citep{pmlr-v202-ball23rlpd}. Within each source, 75\% of the transitions are drawn from steps
within $\rho \le 10$ of their goal (Equation~\ref{eq:rho}), that is, within 1\,cm and $2^\circ$, and the rest
uniformly. Such near-goal transitions are rare in the early online stream, because most steps of an episode are spent on
the approach. They train terminal correction and holding directly. We found that this emphasis also raises the
overall reaching rate, not only the final precision, presumably because values that are accurate near the goal
propagate to the approach through bootstrapping.

\paragraph{Goal relabeling.}
We relabel goals with hindsight experience replay (HER)~\citep{andrychowicz2017hindsight}. Adapting the goal-sampling
distributions of OGBench~\citep{park2025ogbench} to online training, each sampled transition receives its goal from
one of three sources:
\begin{itemize}
\item the \emph{commanded} goal of its episode, which keeps training on the task that the episode attempted,
including failed attempts;
\item a \emph{future} goal, the pose achieved a geometrically distributed number of steps later in the same episode,
with offset $k \sim \mathrm{Geom}(1 - \gamma)$, which turns every trajectory into successful examples for the poses
it passes through and so densifies the reaching signal;
\item the \emph{current} goal, the pose achieved at the transition itself, which supplies examples of being at the
goal and so trains holding. In 30\% of draws it is perturbed by Gaussian noise of 1\,mm and $0.2^\circ$, which places
the goal just around the current pose and trains corrections at the scale of the tolerance.
\end{itemize}
OGBench also uses random dataset states as goals, but we do not. Online transitions use the three sources with
probabilities 0.4, 0.4, and 0.2. Demonstration transitions use 0.5, 0.3, and 0.2, because their actions are directed
at the commanded goal and keeping that goal more often preserves this signal.

\paragraph{Demonstrations.}
An off-policy learner can use demonstrations without changes to the learner. With the mixing above, demonstrations
recorded through the same action interface are added to training and enter the same goal relabeling. When
demonstrations are collected, the budget is best spent where the policy's own exploration produces successes least
often: tasks that require a large reconfiguration in joint space, which the near-goal curriculum of
Appendix~\ref{app:rt_tasks} rarely produces early in training. Our collection protocol follows from this. We record
150 tasks stratified by start-to-goal rotation ($60$--$90^\circ$, $90$--$120^\circ$, and $120$--$180^\circ$) and by
three radial shells of the workspace, each with a joint-space distance $\| q^* - q_0 \|_2 \ge 3$\,rad. The
demonstrations run through the training environment with the same action law and gain schedule, so every recorded
action is an action the policy can take. A scripted controller suffices because each task stores its goal
configuration $q^*$. It moves the reference along the joint-space line to $q^*$ at full step size,
$a_t = \mathrm{clip}\big( (q^* - z_{t-1}) / (\beta_t \eta),\, -1,\, 1 \big)$, until every joint is within 0.002\,rad,
and then applies the integral correction $a_t = \mathrm{clip}\big( 0.5\, (q^* - q_t) / (\beta_t \eta),\, -1,\, 1 \big)$
for the rest of the 150-step episode. The rewards of the
demonstrations are recomputed under the training reward.

\subsection{Training Tasks}
\label{app:rt_tasks}

Each training episode lasts 150 control steps with one goal. Its task comes from one of two sources.

\paragraph{Reverse curriculum.}
Following \citet{pmlr-v78-florensa17reverse}, starts are generated near goals. A goal configuration $q^*$ is drawn from a
precomputed set of reachable, collision-free configurations in the workspace, and the start is $q_0 = q^* + \epsilon$
with $\epsilon \sim \mathcal{N}(0, \sigma_{\mathrm{ep}}^2 I)$. The scale $\sigma_{\mathrm{ep}}$ is drawn uniformly from
$[2^\circ, \sigma_{\max}]$, and $\sigma_{\max}$ is $10^\circ$ in the first half of training and $20^\circ$ in the
second. Starts with $\rho < 10$ are rejected, so every episode includes an approach. Every such task is solvable,
since undoing the perturbation reaches the goal.

\paragraph{Far-task pool.}
A pool of 3{,}000 start--goal pairs provides long motions. Each pair has a start-to-goal rotation of $60$--$170^\circ$
and a joint-space distance $\| q^* - q_0 \|_2 \ge 3$\,rad, and a pool episode draws one pair uniformly. The share of
pool episodes rises from 0 to 25\% at 10\% of training and to 50\% at 20\%. The first episodes therefore stay near
their goals, which teaches terminal regulation first and keeps early motions small. No evaluation start--goal pair
appears in the pool or among the demonstration tasks.

\subsection{Environment and Reward}
\label{app:rt_env}

The hardware environment follows Section~\ref{sec:method} with the values of Table~\ref{tab:rt_params}. The 72-dimensional
observation contains the measured joint state, the end-effector and goal poses, the command-state feedback $d_t$,
the previous action, the keypoint offsets, the coarse and fine error channels, and the running-best features. The
action scale $\eta = 0.03$\,rad and the gain floor $\beta_{\min} = 0.03$ give a smallest reference step of
$\beta_{\min} \eta \approx 0.05^\circ$ per joint, and the reference may lead the measured joints by at most
0.25\,rad.

\paragraph{Relaxed reference bound.}
Equation~\ref{eq:reference-update} clips the reference at the joint bounds. A joint that holds a goal near its bound against
gravity needs a reference beyond the goal configuration, which a hard clip at the bound
forbids. On hardware we relax the clip by $2^\circ$ while the measured joint is still inside the bound. Once the
measured joint reaches the bound, no further outward step is accepted, so the joint itself never exceeds it.

\paragraph{Reward.}
The reward terms are those of Section~\ref{sec:reward}. The shaping potential is capped at 500 tolerance units
instead of 50, so that its dense signal covers the large hardware tasks. Because the shaping term is
potential-based~\citep{ng1999policy}, its contributions along a trajectory telescope, and it does not penalize a
detour that temporarily increases the error. The joint-margin weight $w_m = 5$ keeps the joints away from their
bounds during the approach, where the margin cost applies ($\rho > 10$). Every term is multiplied by the control
period of 0.1\,s.

\subsection{Ablation of Hardware Components}
\label{app:rt_ablation}

We compare PolicyIK with a policy trained without three components of the
recipe, each of which addresses one failure mode: a critic that undervalues
detours through joint space, a far-task pool that under-samples large
joint-space reconfigurations, and goals that cannot be held at a joint bound:
\begin{itemize}
\item \textbf{Latent-alignment critic.} The ablated policy uses FlashSAC's critic (Figure~\ref{fig:rt_arch}a, two
blocks of width 256). Its motivation is the mismatch between pose distance and temporal distance discussed in
Appendix~\ref{app:rt_adapt}.
\item \textbf{Far-task selection by joint-space distance.} The ablated policy draws its far tasks from 150
start--goal pairs selected at random within the same rotation strata, without a joint-space distance criterion. The
joint-space distance measures how much reconfiguration a task needs more directly than the end-effector rotation.
Selecting far tasks by it concentrates the long episodes on reconfigurations, and the larger pool avoids revisiting
a few pairs.
\item \textbf{Relaxed reference bound.} The ablated policy clips the reference at the joint bounds, so a goal whose
holding offset points past a bound cannot be held.
\end{itemize}
Both policies are trained for 120{,}000 online transitions on the same arm and evaluated on the same 60 tasks (Appendix~\ref{app:rt_eval}).

\begin{table}[t]
\centering
\caption{Ablation of the hardware components on 60 physical OpenArm tasks. The \emph{Ablated} policy excludes the latent-alignment critic, far-task selection by joint-space distance, and the relaxed reference bound. Reached and held use
the criteria defined in the main body. Coarse counts tasks whose final-second errors are within 1\,cm and
$2^\circ$, and gross counts tasks that end more than 20\,mm from the goal, both at the 25\,s budget. Bold marks the
better result.}
\label{tab:rt_ablation}
\small
\begin{tabular}{lcccccccc}
\toprule
 & \multicolumn{3}{c}{15\,s budget} & \multicolumn{5}{c}{25\,s budget} \\
\cmidrule(lr){2-4}\cmidrule(lr){5-9}
Policy & Reached & Held 1\,s & Held 5\,s & Reached & Held 1\,s & Held 5\,s & Coarse & Gross \\
\midrule
Ablated & 40 & 36 & \textbf{31} & 46 & 43 & 40 & 52 & 5 \\
PolicyIK & \textbf{52} & \textbf{44} & 27 & \textbf{54} & \textbf{45} & \textbf{43} & \textbf{59} & \textbf{1} \\
\bottomrule
\end{tabular}
\end{table}

Table~\ref{tab:rt_ablation} shows the result. With the three components, PolicyIK reaches 54 instead of 46 tasks
within 25\,s and holds 43 instead of 40 for the final 5\,s. It ends 59 tasks within the coarse tolerance, against
52, and one task more than 20\,mm from the goal, against five, so the components mainly remove failures to approach
the goal. Its median final errors are 0.50\,mm and $0.12^\circ$, against 0.53\,mm and $0.12^\circ$. At 15\,s it
reaches 52 tasks instead of 40, but holds 27 instead of 31 for 5\,s. The additional tasks include long motions that
reach the tolerance too late for a 5\,s hold within 15\,s, and most of them are held for 5\,s by 25\,s.

\subsection{Evaluation Protocol}
\label{app:rt_eval}

We evaluate frozen checkpoints on the 60 full-pose tasks used for the classical controllers
(Appendix~\ref{app:baselines}), so the requested joint-space starts and world-frame goals match across controller
families. Each trial starts at rest from the task's start configuration, and the policy uses its mean
action. We evaluate the final checkpoint after 120{,}000 online transitions, and score the
15\,s and 25\,s budgets on the same trial.

\begin{table}[ht]
\centering
\caption{Reaching and final-window holding counts out of 60 OpenArm tasks, with best results in bold and second-best results underlined.}
\label{tab:hardware-windows}
\setlength{\tabcolsep}{5pt}
\begin{tabular}{lrrrrrr}
\toprule
& \multicolumn{3}{c}{15 s budget} & \multicolumn{3}{c}{25 s budget} \\
Controller & Reached & Held 1 s & Held 5 s & Reached & Held 1 s & Held 5 s \\
\midrule
DLS & 38 & \underline{21} & \underline{6} & \underline{48} & \underline{29} & \underline{17} \\
Resolved-rate QP & \underline{43} & \underline{21} & \underline{6} & \underline{48} & 24 & 16 \\
Kinematic MPC & 29 & 10 & 3 & 46 & 26 & 15 \\
Adapted OpenArm QP & 13 & 8 & 1 & 32 & 16 & 11 \\
\rowcolor{policyikrow} PolicyIK & \textbf{52} & \textbf{44} & \textbf{27} & \textbf{54} & \textbf{45} & \textbf{43} \\
\bottomrule
\end{tabular}
\end{table}

Table~\ref{tab:hardware-windows} shows that PolicyIK keeps its holding advantage under both execution budgets. At 15\,s, it holds
27 of the 60 tasks for the final 5\,s, compared with at most 6 for the classical controllers. Extending the required
final hold from 1\,s to 5\,s retains 61.4\% and 95.6\% of PolicyIK's successes at the two budgets, against
12.5--30.0\% and 57.7--68.8\% for the classical controllers.

\begin{table}[p]
\centering
\caption{Settings for training PolicyIK on hardware, used for both the OpenArm and the xArm7. The FlashSAC column
lists the corresponding settings of FlashSAC~\citep{kim2026flashsac}; a dash marks a
setting that FlashSAC does not specify.}
\label{tab:rt_params}
\small
\begin{tabular}{p{5.0cm}p{4.8cm}p{2.8cm}}
\toprule
Setting & PolicyIK on hardware & FlashSAC \\
\midrule
\multicolumn{3}{l}{\textit{Learner}} \\
Online transitions & 120{,}000 & \\
Batch size & 1{,}024 & 2{,}048 \\
Replay capacity & 120{,}000 (the whole run) & $10^6$ \\
Initial random actions & 200 & -- \\
Updates begin after & 1{,}024 transitions & -- \\
Gradient updates & 2.5 (Appendix~\ref{app:rt_adapt}) & 1 per transition \\
Discount $\gamma$ & 0.98 & 0.99 \\
Target update rate $\tau$ & 0.005 & 0.01 \\
Learning rate (Adam) & $3 \times 10^{-4}$, constant & $3 \times 10^{-4}$, cosine decay to $1.5 \times 10^{-4}$ \\
Actor & 2 blocks, width 128 & 2 blocks, width 128 \\
Critic & latent-alignment critic: $\phi$ 2 blocks, $\psi$ 1 block, width 256 & 2 blocks, width 256 \\
Atoms $n_{\mathrm{atom}}$ on $[G_{\min}, G_{\max}]$ & 201 on $[-5, 5]$ & 101 on $[-5, 5]$ \\
Entropy target $\sigma_{\mathrm{tgt}}$ & 0.30 & 0.15 \\
Initial temperature & 0.01 & 0.01 \\
Actor update period & 2 critic updates & 2 critic updates \\
Mixed precision & off & on \\
\midrule
\multicolumn{3}{l}{\textit{Replay}} \\
Demonstration share of a batch & 25\% & \\
Near-goal share within each source & 75\% from steps with $\rho \le 10$ & \\
Goal sources: commanded, future, current & online 0.4, 0.4, 0.2; demonstrations 0.5, 0.3, 0.2 & \\
Future-goal offset & $\mathrm{Geom}(1 - \gamma)$ steps & \\
Current-goal perturbation & 30\% of draws; 1\,mm, $0.2^\circ$ & \\
Demonstrations & 150 tasks $\times$ 150 steps & \\
\midrule
\multicolumn{3}{l}{\textit{Training tasks}} \\
Episode length & 150 control steps & \\
Start perturbation $\sigma_{\mathrm{ep}}$ & $\mathcal{U}[2^\circ, \sigma_{\max}]$; $\sigma_{\max}$ $10^\circ$, then $20^\circ$ from 50\% & \\
Rejected starts & $\rho < 10$ & \\
Start and goal jitter & $2^\circ$ and $1^\circ$ per joint & \\
Far-task pool & 3{,}000 pairs; rotation $60$--$170^\circ$; $\| q^* - q_0 \|_2 \ge 3$\,rad & \\
Far-task share & 0, 25\% from 10\%, 50\% from 20\% of training & \\
\midrule
\multicolumn{3}{l}{\textit{Environment}} \\
Control rate & 10\,Hz & \\
Observation dimension & 72 & \\
Action scale $\eta$ & 0.03\,rad & \\
Gain floor $\beta_{\min}$ & 0.03 & \\
Gain thresholds $(l_p, u_p)$, $(l_R, u_R)$ & $(0.5, 9.31)$\,mm, $(0.415, 2.328)^\circ$ & \\
Reference lead bound & 0.25\,rad & \\
Reference bound relaxation & $2^\circ$ & \\
Reset between episodes & 4\,s ramp, 1.5\,s settling & \\
\midrule
\multicolumn{3}{l}{\textit{Reward}} \\
Shaping potential $\Phi_t$ & $-\min(500, \rho_t)$ & \\
Motion-cost weights $w_a$, $w_v$ & 0.05, 0.0005 & \\
Joint-margin weight $w_m$ & 5 & \\
Term scaling & control period, 0.1\,s & \\
\bottomrule
\end{tabular}
\end{table}

\clearpage
\section{Hardware Baseline Implementations}
\label{app:baselines}

We compare four Cartesian controllers on the physical OpenArm: damped
least-squares differential inverse kinematics (DLS), bounded resolved-rate
quadratic programming (QP), locally linearized kinematic model-predictive
control (MPC), and the QP controller distributed with the OpenArm control
library (adapted OpenArm QP). All four controllers operate through the same
supervised execution stack. They receive the measured seven-dimensional joint
state, the desired world-frame end-effector pose, and the previous accepted
joint reference, and return the next joint-position reference. None uses an
external pose measurement or inverse dynamics in its Cartesian optimization.
The shared motor layer tracks accepted joint references using PD control with
model-based gravity feedforward.

\paragraph{Shared pose, command, and safety interface.}
Let $q$ denote the measured joint positions, $z$ the previous accepted joint
reference, $\mathbf e=[\mathbf e_p^\top,\mathbf e_R^\top]^\top$ the world-frame position and
rotation-vector error, and $J(q)\in\mathbb{R}^{6\times7}$ the geometric
Jacobian. For DLS, resolved-rate QP, and kinematic MPC, forward kinematics and
$J$ are evaluated from the same MuJoCo model at the measured $q$. Translation
and rotation are balanced using
\begin{equation}
    W=\operatorname{diag}(1,1,1,0.2,0.2,0.2).
\end{equation}
The adapted OpenArm QP uses its separately pinned OpenArm model internally, while
the common model is retained for evaluation. The evaluated end-effector is a
gripper-base frame translated by $102.5$~mm along the source hand-TCP
$z$ axis with no rotational offset; this transform is used consistently for
goal generation, control, and scoring.

Accepted commands are constrained by the intersection of the robot-model and
physical-rig joint limits, a per-joint reference-speed bound of
$0.20$~rad/s, and a command-to-measurement lead bound. The final baseline
profile maintains a $5$~mrad inward command margin from the hard joint limits.

\paragraph{Damped least-squares controller.}
The DLS controller first computes
\begin{equation}
  \dot q_{\mathrm{dls}}
  =(WJ)^\top\!\left[(WJ)(WJ)^\top+\lambda^2I\right]^{-1}kW\mathbf e,
  \qquad k=2,\quad \lambda=0.02.
\end{equation}
It uniformly rescales this velocity when its infinity norm exceeds the joint-speed bound and integrates from $z$.  When that candidate would violate a
joint-position, per-cycle speed, or reference-lead bound, the implementation
does not clip joints independently.  Instead, it resolves the same globally
scaled damped task as a bounded least-squares problem over the feasible joint-velocity box.

\paragraph{Bounded resolved-rate QP.}
The resolved-rate controller directly solves
\begin{equation}
  \min_{\dot q}\;
  \frac{1}{2}\lVert WJ\dot q-kW\mathbf e\rVert_2^2
  +\frac{\lambda^2}{2}\lVert\dot q\rVert_2^2,
  \qquad k=3,\quad \lambda=0.005,
\end{equation}
subject to the joint-position, per-cycle speed, and command-to-measurement lead
bounds before setting $z^+=z+\Delta t\dot q$.  We solve the dense bounded
least-squares form with SciPy's BVLS implementation (exact least-squares
subsolver, tolerance $10^{-10}$, at most 100 iterations).  DLS and QP therefore
share the same local task family but differ under saturation: DLS first applies
a single global speed scale and only then redistributes a bound-violating step,
whereas QP optimizes the unscaled task directly inside the feasible box.

\paragraph{Kinematic MPC.}
The MPC controller replans at every feedback cycle and applies only the first
control of a seven-step horizon ($0.14$~s at the nominal rate).  It freezes the
measured-state Jacobian within each solve and optimizes a sequence of joint-reference velocities using the same $k=3$, $\lambda=0.005$, and $W$ as QP.
The objective combines predicted resolved-rate residuals, velocity damping, a
$0.001$ velocity-smoothness weight, and a unit terminal pose-error weight.
Constraints cover reference and predicted joint positions, the $0.20$~rad/s
speed bound, a $0.052$~rad reference-lead bound, a $0.10$~rad trust region,
and $0.5$~rad/s$^2$ reference acceleration.  A piecewise-linear braking
envelope reserves stopping distance near position and lead limits.  The convex
QP is solved with DAQP~0.7.2 using primal and dual tolerances of $10^{-11}$ and
$10^{-12}$ and an iteration limit of 20,000.

\paragraph{Adapted OpenArm QP.}
The OpenArm baseline uses \texttt{openarm\_control}~0.4.0 with its DAQP
backend. At 50~Hz, the controller performs five internal iterations with an
explicit right-arm velocity limit of $0.20$~rad/s. The configured task
parameters use position and orientation costs of $12$ and $1.5$, damping
$0.1$, Levenberg--Marquardt damping $0.01$, joint and singularity braking,
and zero posture and null-space costs.

Our adapter supplies measured joint positions to the library's
state-dependent constraints, synchronizes the internal task state to the
previous accepted reference, and transforms the world-frame goal into the
library representation. The returned candidate is then projected through the
common hard and operational position bounds, reference-speed bound, and
$0.05$~rad command-to-measurement lead bound. No unconstrained retry or
fallback controller is used.

The evaluated implementation additionally includes a bounded measured-pose
settling loop. Once the commanded pose remains within
$0.1$~mm/$0.001$~rad and the measured pose lies within
$10$~mm/$0.05$~rad for $0.2$~s, the loop adjusts the internal pose target
using the measured residual error. Translational and rotational correction
rates are limited to $1$~mm/s and $0.01$~rad/s, with target-bias limits of
$10$~mm and $0.05$~rad, respectively. This modification compensates residual
command--measurement bias while retaining the upstream reaching controller
and shared command governor.

\begin{table}[t]
\centering
\caption{Classical hardware baselines. All four map the measured joint state and
the goal pose to joint-position references at 50~Hz within the joint limits and
the 0.20~rad/s reference-speed bound. DLS, QP, and MPC weight rotation errors by
0.2 relative to translation.}
\label{tab:hardware-baseline-settings}
\footnotesize
\begin{tabularx}{\linewidth}{@{}l>{\raggedright\arraybackslash}Xl@{}}
\toprule
Controller & Formulation & Parameters \\
\midrule
DLS & Damped least squares with global speed scaling & $k=2$, $\lambda=0.02$ \\
Resolved-rate QP & Damped least squares within the joint bounds & $k=3$, $\lambda=0.005$ \\
Kinematic MPC & Seven-step receding horizon (0.14~s) & $k=3$, $\lambda=0.005$ \\
Adapted OpenArm QP & OpenArm QP with measured-pose settling & Costs 12/1.5, damping 0.1 \\
\bottomrule
\end{tabularx}
\end{table}

\paragraph{Parameter selection.}
Controller settings were selected during hardware development.
The controller descriptions and Table~\ref{tab:hardware-baseline-settings}
specify the final settings.

\paragraph{Classical-controller evaluation protocol.}
The evaluation set contains 60 full-pose start--goal tasks, with 20 tasks in
each of three translational-displacement bands: medium
($>2$ to $\leq5$~cm), long ($>5$ to $\leq10$~cm), and extended
($>10$~cm). The realized displacement ranges are 2.13--4.81~cm,
5.25--9.71~cm, and 10.79--32.93~cm, respectively. Each trial runs for 40~s after a synchronized transition to the specified
start configuration.
The 15~s and 25~s budgets are scored on the first 15~s and 25~s of each
trial.

\section{Cross-Embodiment Evaluation on xArm7}
\label{app:xarm}
We further evaluate PolicyIK on a commercial xArm7. Its wrist joints have a wider range of motion than those of OpenArm, which makes full-pose tasks easier on this arm. Both simulation-trained and hardware-trained policies achieve 60/60 successes on standard reaching and 60/60 on large rotations (Table~\ref{tab:xarm}).

\begin{table*}[!ht]
\centering
\small
\caption{Physical xArm7 evaluation on two sets of 60 tasks each. Reach-time medians use reached tasks.}
\label{tab:xarm}
\begin{tabular}{lrrrr}
\toprule
Controller & Success & Reach (s) & Position (mm) & Orientation ($^\circ$) \\
\midrule
\multicolumn{5}{c}{Standard reaching} \\
DLS & 60/60 & 2.9 & 0.074 & 0.015 \\
QP & 60/60 & 2.0 & 0.094 & 0.018 \\
MPC & 60/60 & 2.3 & 0.085 & 0.016 \\
xArm built-in IK & 59/60 & 1.8 & 0.004 & 0.001 \\
\rowcolor{policyikrow} PolicyIK (sim-trained) & 60/60 & 2.2 & 0.169 & 0.034 \\
\rowcolor{policyikrow} PolicyIK (hardware-trained) & 60/60 & 2.0 & 0.155 & 0.040 \\
\midrule
\multicolumn{5}{c}{Large rotations} \\
DLS & 59/60 & 5.4 & 0.074 & 0.016 \\
QP & 59/60 & 4.2 & 0.087 & 0.020 \\
MPC & 59/60 & 5.0 & 0.086 & 0.018 \\
xArm built-in IK & 45/60 & 4.1 & 0.004 & 0.001 \\
\rowcolor{policyikrow} PolicyIK (sim-trained) & 60/60 & 4.5 & 0.097 & 0.028 \\
\rowcolor{policyikrow} PolicyIK (hardware-trained) & 60/60 & 4.4 & 0.164 & 0.044 \\
\bottomrule
\end{tabular}
\end{table*}

\paragraph{Setup.}
Each task set contains 60 fixed start--goal tasks, evaluated once per controller for 40~s. Large-rotation tasks cover start-to-goal orientation changes of $30$--$60^\circ$, $60$--$90^\circ$, and above $90^\circ$. Success uses the holding criterion of Section~\ref{sec:hardware}: both errors remain within 1~mm and $0.2^\circ$ throughout seconds 20--25 of each trial, the final 5~s of its 25~s budget. Reach times and pose errors are computed as in Section~\ref{sec:hardware}. All controllers share a 0.20~rad/s joint-speed limit. DLS, QP, and MPC use the gains, damping, and horizon of Appendix~\ref{app:baselines}. xArm7 training adopts the same PolicyIK hyperparameters as for training the OpenArm in Table~\ref{tab:rt_params}.

\paragraph{Training.}
We train separate xArm7 policies with the FlashSAC recipe of Section~\ref{sec:real-implementation}, both in simulation and directly on the physical robot. Hardware training starts without simulation weights and collects approximately 120k online interactive transitions, supplemented by demonstrations collected on the arm.

\paragraph{Results.}
PolicyIK exceeds all evaluated baselines in success rate on large rotations and matches the best of them on standard reaching. On large rotations, both training routes achieve 60/60 successes, compared with 59/60 for DLS, QP, and MPC and 45/60 for the built-in IK. On standard reaching, both achieve 60/60, matching DLS, QP, and MPC. These results demonstrate precise physical control on a second robot through both simulation training and direct hardware learning.

\end{document}